\documentclass[11pt,a4paper]{article}

\usepackage[margin=1in]{geometry}
\usepackage[numbers,sort&compress]{natbib}
\usepackage{doi}
\usepackage{booktabs}
\usepackage{siunitx}
\usepackage{authblk}
\usepackage{amsmath}
\usepackage{graphicx}
\usepackage{amssymb}
\usepackage{booktabs,longtable,tabularx,array}
\usepackage{enumitem}
\usepackage{caption}
\usepackage{makecell}
\usepackage{multirow}
\usepackage[version=4]{mhchem}
\usepackage{xcolor}
\usepackage{pifont}
\usepackage{multirow}
\usepackage[ruled]{algorithm2e}

\usepackage{parskip}
\usepackage{microtype}
\usepackage[T1]{fontenc}
\usepackage{lmodern}

\title{
\LARGE \textbf{Beyond DFT electronic band structures via dual-space orbital learning}
}

\author[*,1,3]{Minhyuk Kang}
\author[*,2]{Beom Seok Kang}
\author[1,4]{Nikolai Argatoff}
\author[2]{Miguel Nouman}
\author[1]{Danish Khan}
\author[2,$\dagger$]{William A. Goddard III}
\author[1,$\dagger$]{Anima Anandkumar}

\affil[1]{Division of Engineering and Applied Science, California Institute of Technology}
\affil[2]{Division of Chemistry and Chemical Engineering, California Institute of Technology}
\affil[3]{Department of Chemical Engineering, Imperial College London}
\affil[4]{Department of Mathematics, ETH Z\"urich}
\affil[*]{Equal Contribution}
\affil[$\dagger$]{Corresponding author(s). E-mail(s): wag@caltech.edu; anima@caltech.edu}

\date{}

\hypersetup{
  hidelinks,
  pdfusetitle
}

\newcommand{\ourmodel}{O-kRystal}

\newcommand{\gzwz}{\ensuremath{G_0W_0}}
\newcommand{\rs}{\mathbf{R}}
\newcommand{\ks}{\mathbf{k}}
\newcommand{\eref}{E_{\mathrm{mid}}}
\newcommand{\ereftext}{band-edge midpoint}

\begin{document}
\maketitle

\section*{Abstract}

Accurate electronic band structures are essential for predicting material properties, yet high-fidelity calculations remain computationally demanding.
Here we introduce \ourmodel{}, a dual-space orbital-learning framework that uses a lower-cost Perdew--Burke--Ernzerhof (PBE) calculation both as an energy baseline and as a source of physically structured orbital information, combining band energies within real- and reciprocal-space orbital matrices to predict higher-level corrections to individual electronic states as a function of crystal momentum $\ks$ and spin.
Using a $\pm5$~eV window around the \ereftext{}, \ourmodel{} predicts Heyd--Scuseria--Ernzerhof (HSE06)-level band energies with a mean absolute error of $0.171\pm0.009$~eV across nearly 6,000 three-dimensional materials, spanning insulating/metallic and magnetic/non-magnetic electronic structures.
On a separately held-out benchmark of six strongly correlated materials, it reduces the corresponding PBE band-energy MAE by 68.7\%, while resolving orbital-dependent changes in the relative placement of transition-metal $d$ and ligand $p$ states.
Even with supervision only at high-symmetry points, the model maintains comparable accuracy at unseen path-interior $\ks$ points and reconstructs HSE06 densities of states from dense Brillouin-zone predictions.
Finally, training \ourmodel{} for \gzwz{} quasiparticle energies of two-dimensional materials demonstrates applicability across levels of theory, establishing orbital learning as a general strategy for higher-fidelity $\ks$-resolved electronic-structure prediction beyond density functional theory (DFT) calculations.

\section{Introduction}

Electronic band structure provides a momentum ($\ks$)-resolved description of the allowed electronic states of a crystalline material \citep{AshcroftMermin1976, slater1954blochsum}.
While scalar quantities such as the band gap compress this electronic structure into individual descriptors, the full dispersion retains the energies, curvature, crossings, and spin dependence of individual states.
These features govern carrier dynamics and are central to electronic and thermal transport, where the relative energies and shapes of the bands strongly influence material response \citep{zhou2026electronictransport, whalley2019optical_transport, pei2011bandconvergence, park2021bandconvergence}.
Consequently, accurate band dispersions provide microscopic design information for screening and engineering materials for photovoltaic, electronic, and thermoelectric devices \citep{Yu2012PV, Gorai2017Thermoelectrics}.
Momentum-resolved electronic structure is also particularly important in correlated, topological, and other quantum materials, where changes in the dispersion can define qualitatively different electronic phases, including metallic, insulating, and topological states, and can be probed experimentally by angle-resolved photoemission spectroscopy \citep{zhang2009topological_mat, yang2018arpes, lv2019arpes, xian2023bandmapping}.
Predicting the full band structure therefore preserves significantly more of the underlying electronic-state information than learning a predefined set of scalar properties.

First-principles electronic-structure methods provide a systematic framework for calculating band structures, but improved descriptions of electronic states generally come at greater computational cost \citep{Hohenberg1964HK,Kohn1965KS}.
Density-functional theory (DFT) with semilocal exchange--correlation functionals, such as Perdew--Burke--Ernzerhof (PBE), is efficient for large-scale materials calculations \citep{Perdew1996PBE}, but typically underestimates band gaps because of well-known limitations associated with delocalization error and the absence of the exchange-correlation derivative discontinuity \citep{perdew1983pbe_bad1,mori2008pbe_bad2,crowley2016pbe_bad3}.
Hybrid functionals such as Heyd--Scuseria--Ernzerhof (HSE06) incorporate screened nonlocal exact exchange, reducing delocalization errors and systematically improving band gaps and electronic state energies \citep{Heyd2003HSE, Krukau2006HSE06, liu2024hse_good}.
Many-body approaches, such as the Green's function based \textit{GW} approximation, use the electronic self-energy to obtain quasiparticle addition and removal energies that are more directly connected to spectroscopic observables \citep{Hedin1965GW,golze2019gw_review}.
Both approaches are substantially more expensive than semilocal DFT: conventional hybrid-functional and \gzwz{} implementations can exhibit asymptotic scaling of approximately $\mathcal{O}(N^4)$ with system size, compared with $\mathcal{O}(N^3)$ for Kohn--Sham DFT.
Additionally, their cost can also grow more steeply with the number of sampled $\ks$-points, making many calculations prohibitively expensive and sometimes infeasible \citep{Redies2022HybridScaling, Grossmann2024GWConvergence}.
This cost--accuracy trade-off motivates methods that retain the electronic information already available from an inexpensive baseline calculation while efficiently predicting the corresponding target-theory band structure.
Importantly, both HSE06 and $GW$ remain closely connected to a lower-level electronic-structure starting point \citep{Krukau2006HSE06,golze2019gw_review}.
HSE06 operates within a generalized Kohn--Sham framework, while $GW$ calculations are typically constructed from mean-field orbitals and eigenvalues.
This makes the orbital information obtained from an inexpensive DFT calculation a natural, physically structured representation for learning corrections to higher-level electronic structure.

Machine learning has increasingly been used to accelerate electronic-structure prediction, from scalar material properties to complete state-resolved descriptions \citep{Tang2025DeepElectronicStructure}.
Structure-based machine-learning models can efficiently predict quantities such as band gaps or densities of states directly from crystal structures \citep{Xie2018CGCNN, Choudhary2021ALIGNN, Zhuo2018BandGaps}, while direct structure-to-band approaches predict the electronic dispersion without an intermediate electronic-structure calculation \citep{Gong2024Bandformer}. 
At a more advanced level, equivariant neural networks can learn electronic Hamiltonians from atomic structure and thereby recover band structures through a subsequent diagonalization \citep{Li2022DeepH, gong2023deeph_e3, li2023magneticdeeph, tang2024deeph_hybrid, gong2024planewave}. 
Matrix-supervised Hamiltonian-learning approaches require target-theory labels in a specified basis.
Learning band-energy corrections instead uses eigenvalue targets without requiring target-theory matrix elements.
Recently, neural operators have alternatively targeted the Kohn--Sham potential-to-density map to accelerate self-consistent calculations \citep{Khan2026KSMap}. 

An intermediate strategy is to retain the electronic structure from a lower-cost calculation and learn its correction to a higher level of theory.
This is particularly attractive because state-resolved band and quasiparticle energies are increasingly available in electronic-structure databases and are more directly connected to spectroscopic observables than basis-dependent target-theory Hamiltonians \citep{golze2019gw_review,yang2018arpes,lv2019arpes, nair2025hse06_dataset,Haastrup2018C2DB,gjerding2021c2db}.
Existing band-structure learning approaches, however, remain restricted along different conditions.
\emph{Bandformer} directly predicts DFT-level band dispersions from crystal structure on a large materials dataset, but is demonstrated for non-magnetic materials using a fixed six-band representation near the Fermi level \citep{Gong2024Bandformer}.
Adhikari \textit{et al.} instead learned PBE-to-HSE06 band-energy corrections for 337 bulk semiconducting and insulating compounds, using states from one sampled $\ks$-point per material \citep{Adhikari2023DeltaHSE}.
Kn{\o}sgaard and Thygesen learned PBE-to-\gzwz{} quasiparticle corrections for 286 non-magnetic two-dimensional semiconductors over a fixed manifold of eight valence and four conduction bands \citep{knosgaard2022gw_ml}.

These studies leave an important gap: higher-fidelity electronic band-structure prediction has not yet been demonstrated at scale across chemically diverse materials while simultaneously covering metals and insulators, explicitly resolving spin, and accommodating different numbers of bands across materials.
To our knowledge, no previous framework combines these capabilities within a single state-resolved electronic band-structure learning approach.

Orbital learning offers a natural way to use the information already contained in a lower-cost electronic-structure calculation. 
Rather than predicting target-theory properties from atomic structure alone, OrbNet and its subsequent variants use orbital-based quantities from inexpensive quantum-mechanical calculations to generate input features, matrices representing the electronic-structure (Fock, density, overlap, core-Hamiltonian), which have demonstrated high accuracy and data efficiency across molecular systems \citep{qiao2020orbnet, christensen2021orbnet_denali, qiao2022orbnet_equi, kang2025orbitall}.
These developments show that low-cost orbital information can provide a highly informative and data-efficient representation for learning higher-fidelity quantum-mechanical properties.

Extending orbital learning from molecules to crystalline materials requires an explicit treatment of crystal momentum $\ks$, which labels Bloch states in a periodic solid \citep{AshcroftMermin1976,slater1954blochsum}.
In a localized atomic-orbital basis, electronic operators can be represented either by real-space matrix elements connecting orbitals across lattice translations or by their reciprocal-space Bloch transforms at each $\ks$.
Real-space emphasizes localized orbital interactions, whereas reciprocal-space exposes their momentum dependence and the resulting electronic dispersion \citep{qi2026aircpp}.
We combine both views in \textit{dual-space orbital learning}, allowing the model to jointly encode local electronic interactions and $\ks$-resolved band information.

We introduce \ourmodel{}, a crystal-momentum ($\ks$)-resolved orbital-learning framework for higher-level electronic band structures from a lower-cost PBE calculation.
The PBE calculation provides the baseline band structure through its band energies, while its real- and reciprocal-space orbital matrices offer complementary representations of the underlying electronic state.
For the primary PBE-to-HSE06 task, the model is trained on the difference between matched HSE06 and PBE band energies at high-symmetry $\ks$-points.
\ourmodel{} learns how the energy of each electronic state changes from PBE baseline across band index, crystal momentum, and spin, and adds this predicted correction to the corresponding PBE energy to obtain the target-level band structure.
Because the model predicts electronic states individually, it can accommodate different numbers of valence and conduction bands across materials rather than requiring a fixed-size band representation.

We evaluate the framework on nearly 6,000 chemically diverse three-dimensional materials spanning insulating, metallic, and spin-polarized electronic structures \citep{nair2025hse06_dataset}.
For PBE-to-HSE06 prediction, \ourmodel{} achieves a band-energy MAE of $0.171\pm0.009$~eV for electronic states within $\pm5$~eV of the \ereftext{}.
Even though training uses only high-symmetry $\ks$-points, the model generalizes well to unseen $\ks$-points throughout the Brillouin zone, enabling accurate reconstruction of HSE06 densities of states and derived electronic properties.
\ourmodel{} also recovers the target-theory electronic structure even when the PBE baseline gives the wrong qualitative picture, including materials that are metallic at the PBE level but insulating at HSE06, while retaining systems that remain metallic.
On six separately held-out strongly correlated materials, the model reduces the band-energy MAE over densely sampled Brillouin-zone states within the 5~eV window from 1.532~eV with PBE to 0.479~eV, a 68.7\% reduction.
For the canonical Mott insulator AFM-II MnO, orbital-projected spectra further reveal distinct corrections to Mn $3d$ and O $2p$ state energies, demonstrating behavior beyond a uniform shift of the band edges.
The predicted band structures further improve the predictions of experimentally validated downstream properties, including fundamental band gaps and effective masses.
Finally, when trained separately on 286 two-dimensional semiconductors, the same framework achieves a quasiparticle-energy MAE of $0.126\pm0.017$~eV for PBE-to-\gzwz{} prediction \citep{Haastrup2018C2DB, gjerding2021c2db}.
Together, these results establish \ourmodel{}, to our knowledge, as the first broadly applicable framework for beyond-DFT electronic band-structure prediction across chemically and electronically diverse crystalline materials, opening a route toward scalable high-fidelity electronic-structure prediction across materials and levels of theory.

\section{Results}

\begin{figure}[ht!]
    \centering
    \includegraphics[width=\linewidth]{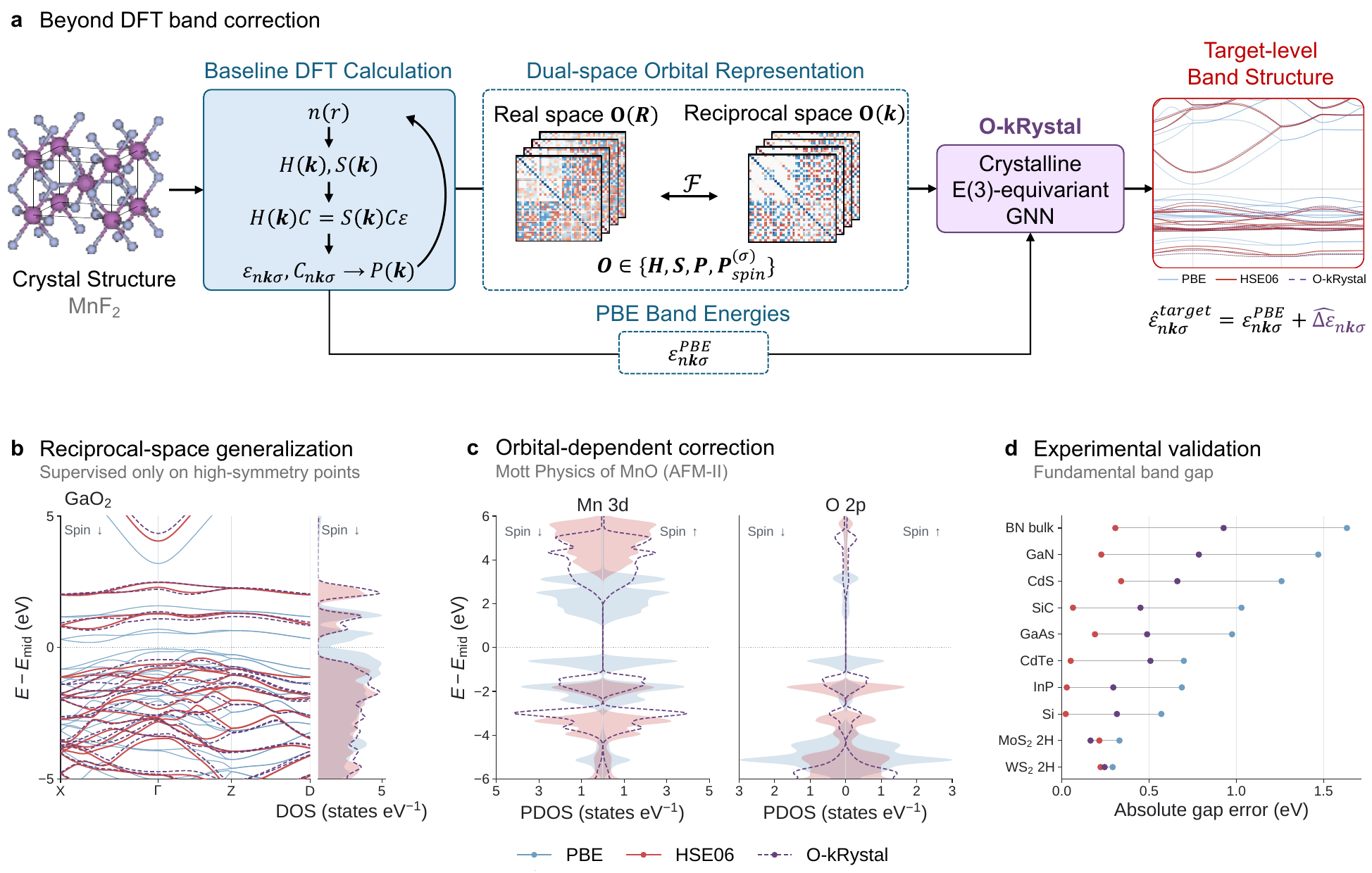}
    \caption{\textbf{Dual-space orbital learning and representative applications.}
    \textbf{a,}
    Starting from a crystal structure, a DFT calculation supplies baseline band energies and orbital matrices, including the Hamiltonian $\mathbf{H}$, overlap $\mathbf{S}$, density $\mathbf{P}$, and spin-density matrices $\mathbf{P}^{(\sigma)}_{spin}$, represented in real ($\rs$)- and reciprocal ($\ks$)-space.
    The dual-space orbital representations are processed by a periodic E(3)-equivariant graph neural network to predict state-dependent energy corrections.
    Target-theory (e.g., HSE06 or \gzwz{}) band energies are reconstructed as $\hat{\varepsilon}^{\mathrm{target}}_{n\ks\sigma} 
    =
    \varepsilon^{\mathrm{base}}_{n\ks\sigma}
    +
    \widehat{\Delta\varepsilon}_{n\ks\sigma}$.
    \textbf{b,}
    Band dispersion and density of states for the displayed spin-down channel of GaO$_2$, comparing PBE, HSE06, and \ourmodel{}.
    For the three-dimensional PBE-to-HSE06 task, supervision is restricted to high-symmetry points.
    The model generalizes to unseen path-interior points, while the density of states is reconstructed from predictions on a dense Brillouin-zone mesh.
    \textbf{c,}
    Spin-resolved Mn $3d$ and O $2p$ projected densities of states for MnO with AFM-II ordering along [111].
    The \ourmodel{} reconstruction combines predicted HSE06 state energies with fixed PBE Mulliken projection weights, illustrating orbital-dependent changes in state-energy placement while retaining the baseline orbital character.
    Resolving the relative placement of transition-metal $d$ and ligand $p$ states is particularly important in strongly correlated oxides, where these energy scales are central to Mott and charge-transfer electronic behavior.
    \textbf{d,}
    Absolute band-gap errors relative to experiment for ten independently evaluated bulk materials from the mstar60 benchmark \citep{laurien2022mstar60_dataset}, compared across PBE, HSE06, and \ourmodel{}
    }
    \label{fig:overview}
\end{figure}

\subsection{The \ourmodel{} Framework}
\label{sec:orbnet_crystal_model}

\ourmodel{} extends orbital learning to periodic electronic structures by formulating target-theory band prediction as a correction to a lower-cost baseline calculation (Fig.~\ref{fig:overview}a). Given baseline band energies $\varepsilon^{\mathrm{base}}_{n\ks\sigma}$, the target energy is reconstructed as
\begin{equation}
\widehat{\varepsilon}^{\mathrm{target}}_{n\ks\sigma}
=
\varepsilon^{\mathrm{base}}_{n\ks\sigma}
+
\widehat{\Delta\varepsilon}_{n\ks\sigma},
\label{eq:delta_learning}
\end{equation}
where $n$, $\ks$, and $\sigma$ denote the band index, crystal momentum, and spin channel, respectively, and the hat denotes a model prediction. In this work, PBE serves as the principal baseline, while HSE06 and \gzwz{} define the target theories \citep{Perdew1996PBE, Heyd2003HSE, Krukau2006HSE06, Hedin1965GW}. This $\Delta$-learning formulation preserves the dispersion already captured by the baseline calculation and focuses the learning problem on the state- and crystal-momentum-dependent correction between electronic-structure methods \citep{ramakrishnan2015delta_learning}.

\paragraph{Orbital representation}

Beyond the baseline eigenvalues, the underlying PBE calculation provides orbital-resolved information about the converged electronic state. \ourmodel{} represents the baseline electronic state through the Kohn--Sham Hamiltonian $\mathbf{H}$, overlap matrix $\mathbf{S}$, density matrix $\mathbf{P}$, and spin-density matrix
$\mathbf{P}^{(\sigma)}_{\mathrm{spin}}
=
\mathbf{P}_{\sigma}-\mathbf{P}_{\bar{\sigma}}$,
where $\sigma$ denotes the spin channel and $\bar{\sigma}$ the opposite spin channel.
On-site matrix blocks describe the electronic environment associated with individual atoms, while off-diagonal blocks encode interactions between atomic orbitals on neighboring sites (Fig.~\ref{fig:overview}b).
Overall, \ourmodel{} uses the lower-cost calculation not only as an energy baseline, but also as a physically structured representation of the material-specific electronic state.

\paragraph{Dual-space periodic representation}

For periodic systems, the same orbital-based quantities can be represented in real- and reciprocal-space.
The real-space representation exposes localized and sparse atom-centered interactions, whereas the reciprocal-space representation naturally provides an explicit $\ks$-resolved description of the corresponding electronic structure (Fig.~\ref{fig:overview}b).
\ourmodel{} processes these complementary representations with a periodic $\mathrm{E}(3)$-equivariant graph neural network.
Real-space on-site blocks define atomic features and pairwise blocks enter equivariant message passing, while Bloch-phase information and reciprocal-space orbital features introduce explicit dependence on the queried crystal momentum.
This formulation enables evaluation at crystal momenta that were not used as supervised training targets.
Detailed definitions of the orbital representation and model architecture are provided in Methods~\ref{sec:methods_representation} and \ref{sec:methods_architecture}.

\subsection{Target-theory band structure prediction}

\begin{figure}[ht!]
    \centering
    \includegraphics[width=\linewidth]{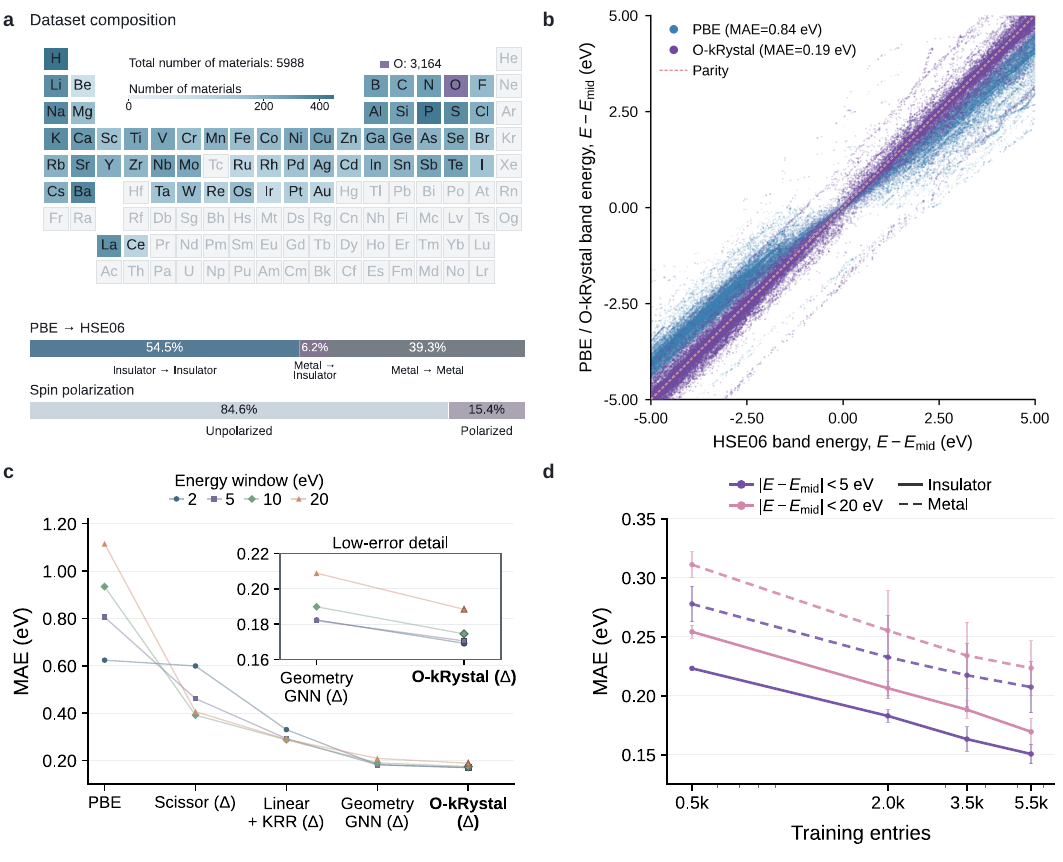}
    \caption{\textbf{Large-scale prediction of HSE06 band structures with \ourmodel{}.}
    \textbf{a,}
    Composition of the three-dimensional materials dataset containing 5,988 chemically diverse parent materials \citep{nair2025hse06_dataset}.
    Element counts indicate the number of materials containing each element.
    Horizontal bars summarize PBE-to-HSE06 electronic-character classes and spin polarization.
    \textbf{b,} 
    PBE and \ourmodel{} band energies compared with HSE06 for 575,487 matched test-state pairs within the PBE-defined 5~eV energy window for a representative held-out test split.
    Energies are independently referenced to the \ereftext{} of each spectrum.
    The dashed line denotes parity, and the reported MAEs are state weighted.
    \textbf{c,}
    Entry-balanced band-energy MAEs for uncorrected PBE, scissor correction, Linear+KRR, Geometry GNN, and \ourmodel{} over energy windows of 2, 5, 10, and 20~eV.
    $\Delta$ identifies the neural models trained directly on state-resolved PBE-to-HSE06 energy corrections.
    \textbf{d,} 
    Learning curves for \ourmodel{} as a function of training-set size, shown separately for insulating and metallic HSE06 entries within the 5 and 20~eV windows.
    Values and error bars in \textbf{c} and \textbf{d} denote the mean and sample standard deviation across three independently trained models with parent-grouped splits.
    Band-energy evaluations in \textbf{b--d} use the supervised high-symmetry $\ks$-points.
    }
    \label{fig:results}
\end{figure}

\paragraph{Dataset and evaluation}

We first evaluate \ourmodel{} on PBE-to-HSE06 band-structure prediction using 6,900 band-structure entries from 5,988 parent materials, spanning broad elemental and electronic-structure diversity, including insulating and metallic as well as magnetic and nonmagnetic materials (Fig.~\ref{fig:results}a) \citep{nair2025hse06_dataset}. The primary benchmark evaluates individual band energies at the supervised high-symmetry $\ks$-points over progressively wider energy windows around the \ereftext{}. Dataset construction, splitting, and evaluation protocols are described in Methods~\ref{sec:methods_data}.

\paragraph{State-resolved prediction accuracy}

\ourmodel{} substantially reduces the state-resolved error of the PBE baseline across both occupied and unoccupied bands (Fig.~\ref{fig:results}b). Evaluated over 575,487 state pairs within $|E-\eref|<$ 5~eV from the representative model shown in Fig.~\ref{fig:results}b, the state-weighted MAE decreases from 0.84~eV for PBE to 0.19~eV with \ourmodel{}. Across three separately trained models using distinct data splits, \ourmodel{} achieves mean entry-balanced band-energy MAEs of $0.169\pm0.008$~eV, $0.171\pm0.009$~eV, $0.174\pm0.010$~eV, and $0.188\pm0.012$~eV within $|E-\eref|<$ 2, 5, 10, and 20~eV, respectively (Fig.~\ref{fig:results}c). The relatively small increase in error as the evaluation window expands indicates that the learned correction is not confined to band-edge energies, but remains effective over a broad portion of the electronic spectrum.

\paragraph{Comparison with correction baselines}

A common approximation to target-level band structures is to apply a simple correction to the baseline PBE energies. Therefore, we compare \ourmodel{} with multiple PBE-to-HSE06 correction strategies, including a rigid scissor correction, a combined Linear+KRR model, and a geometry-based graph neural network trained using the same material splits and prediction targets (Methods~\ref{sec:methods_baseline}; Fig.~\ref{fig:results}c). The uncorrected PBE MAE increases from 0.624~eV within 2~eV of the \ereftext{} to 1.115~eV within 20~eV. The scissor and Linear+KRR corrections reduce the error, with Linear+KRR reaching MAEs of 0.29--0.33~eV. Incorporating crystal geometry through the neural baseline further lowers these errors to 0.18--0.21~eV. \ourmodel{} achieves the lowest errors across all four windows, reaching 0.17--0.19~eV. Relative to the matched Geometry GNN, the reduction increases from 7.3\% within 2~eV to 9.8\% within 20~eV. These results show that the geometry-based model supplied with the PBE band baseline captures a substantial component of the PBE-to-HSE06 correction on the broad test distribution, while explicit orbital information provides a consistent additional improvement across the electronic spectrum. Learning-curve ablations further show that real-space orbital features improve upon the geometry-based model, with explicit reciprocal-space orbital matrices providing an additional advantage that becomes clearer with increasing training-set size and window range (Supplementary Fig.~\ref{fig:s1}).

\paragraph{Learning across electronic regimes}

Prediction accuracy improves systematically as the amount of training data increases across both insulating and metallic systems (Fig.~\ref{fig:results}d). Within $|E-\eref|<5$~eV, the MAE for insulating materials decreases from $0.223\pm0.001$~eV with 0.5k training entries to $0.151\pm0.008$~eV with the full 5.5k training set, while the corresponding metallic error decreases from $0.278\pm0.015$~eV to $0.207\pm0.022$~eV. A similar trend is observed over the broader $|E-\eref|<$ 20~eV window, where the insulating and metallic errors decrease from $0.254\pm0.005$~eV and $0.311\pm0.011$~eV to $0.169\pm0.011$~eV and $0.223\pm0.023$~eV, respectively. Metallic systems remain consistently more challenging across all training-set sizes, reflecting the greater complexity of partially occupied and near-degenerate electronic states. Nevertheless, both regimes continue to improve with additional training data, with no clear performance plateau over the range examined.

Overall, these results establish accurate PBE-to-HSE06 band correction across diverse materials, electronic regimes, and broad energy ranges at the sampled high-symmetry points. We next examine whether the learned $\ks$ dependence extends beyond these reciprocal-space positions to unseen points along band paths and subsequently to off-path points throughout the Brillouin zone.

\subsection{Generalization across the Brillouin Zone}
\label{sec:bz_generalization}

\begin{figure}[ht!]
    \centering
    \includegraphics[width=\linewidth]{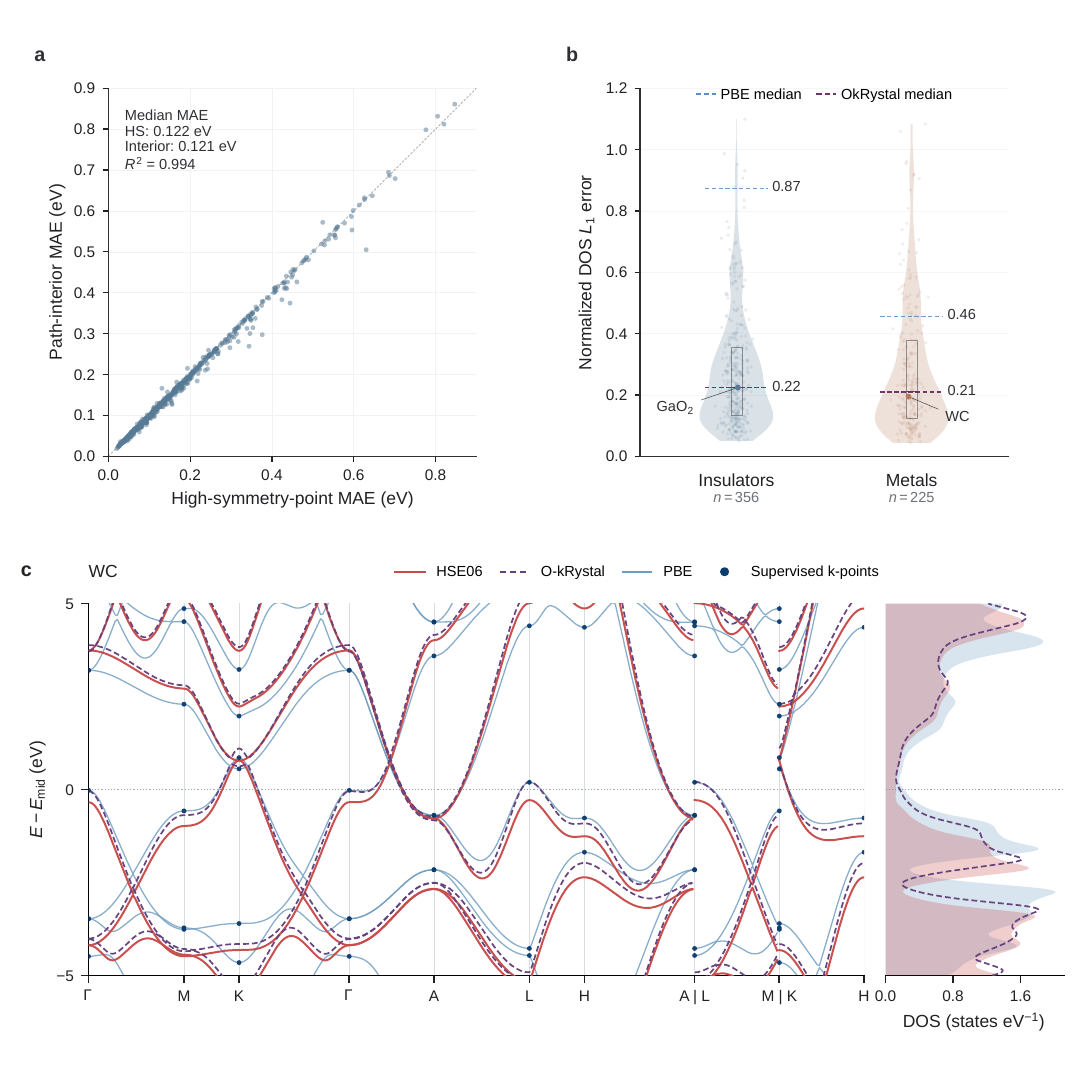}
    \caption{\textbf{Generalization between and beyond high-symmetry $\ks$-points.}
    \textbf{a,}
    Comparison of high-symmetry-point and path-interior band-energy MAEs for 693 held-out band-structure entries within the 5~eV window.
    Each point represents one entry, including separate spin channels where applicable.
    Median MAEs are 0.122~eV and 0.121~eV, respectively, and the dashed line denotes parity.
    \textbf{b,}
    Distribution of normalized density-of-states (DOS) reconstruction errors for 356 insulating and 225 metallic parent materials.
    Spectra are reconstructed from dense off-path predictions and compared with HSE06 over $|E-\eref|<$5~eV using the normalized $L_1$ metric defined in Methods ~\ref{sec:methods_evaluation}.
    Dashed horizontal markers indicate the PBE and \ourmodel{} medians.
    GaO$_2$ and WC identify the examples displayed in Figs.~\ref{fig:overview}b and \ref{fig:generalization}c, respectively.
    \textbf{c,}
    Band structure and density of states of WC, comparing HSE06, \ourmodel{}, and PBE.
    Marked high-symmetry points show the sampling convention used for supervision.
    Intermediate band-path points and the dense mesh used for DOS reconstruction are evaluated without target-theory information at inference.
    }
    \label{fig:generalization}
\end{figure}

Accurate prediction at high-symmetry points alone is insufficient for reconstructing the full electronic dispersion, as many electronic properties depend on states between and away from conventional high-symmetry paths. Thus, we examine whether the $\ks$ dependence learned by \ourmodel{} extends to reciprocal-space positions that are not used as training targets. We evaluate reciprocal-space generalization in two settings: unseen points along the high-symmetry paths and off-path points throughout the three-dimensional Brillouin zone (Fig.~\ref{fig:generalization}).

\paragraph{Prediction at unseen path-interior points}

\ourmodel{} is evaluated at previously unseen $\ks$-points between the high-symmetry points used during training (Fig.~\ref{fig:generalization}a). Across 693 band-structure entries, the median band-energy MAE remains essentially unchanged, from 0.122~eV at the supervised high-symmetry points to 0.121~eV at unseen path-interior points. This negligible degradation shows that the learned correction is not restricted to the discrete reciprocal-space positions encountered during supervision and motivates evaluation beyond the high-symmetry paths. A detailed analysis of the dependence on reciprocal-space distance, relative path position, and segment length is provided in Supplementary Fig.~\ref{fig:s2}

\paragraph{Off-path prediction across the Brillouin zone}

We next consider generalization at off-path $\ks$-points throughout the Brillouin zone. Because the available reference data do not provide consistently matched PBE and HSE06 band states at arbitrary off-path $\ks$-points, we evaluate whether predictions across a dense reciprocal-space grid collectively reconstruct the HSE06 density of states (DOS), which depends on electronic states sampled throughout the Brillouin zone (Fig.~\ref{fig:generalization}b). The reconstruction error is quantified using area-normalized $L_1$ difference from the HSE06 DOS over $\eref\pm5$~eV, as defined in Methods~\ref{sec:methods_evaluation}. Across 356 insulating and 225 metallic materials, \ourmodel{} achieves median DOS errors of 0.224 and 0.210, respectively, reduced from 0.874 and 0.457 for the corresponding PBE electronic structures. 

The representative GaO$_2$ and WC examples, selected near the median \ourmodel{} DOS error of the respective electronic classes, illustrate both forms of reciprocal-space generalization (Figs.~\ref{fig:overview}b and \ref{fig:generalization}c). At unseen path-interior points, \ourmodel{} follows the HSE06 band dispersion substantially more closely than PBE, while dense off-path predictions reproduce the principal features of the corresponding HSE06 DOS. Together, these results show that the learned state corrections remain accurate between the supervised high-symmetry points and support spectral reconstruction from predictions at previously unseen off-path locations throughout the Brillouin zone.

\subsection{Electronic-character recovery and strongly correlated materials}

\begin{figure}[ht!]
    \centering
    \includegraphics[width=\linewidth]{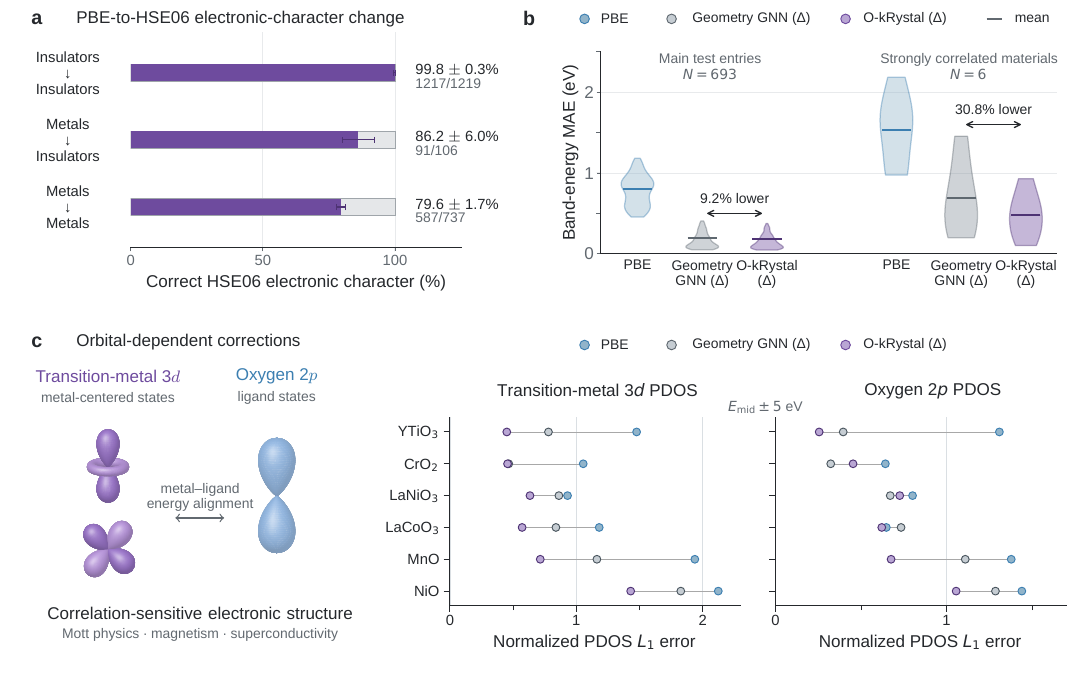}
    \caption{\textbf{Electronic-character recovery and orbital-dependent corrections in strongly correlated materials.}
    \textbf{a,}
    Recovery of HSE06 electronic character for insulator-to-insulator, metal-to-insulator, and metal-to-metal systems.
    Bars show the fraction for which \ourmodel{} predicts the correct HSE06 metallic or insulating character across three independently trained models.
    Adjacent fractions report the number of correct predictions and total entries pooled across the corresponding test evaluations.
    \textbf{b,} 
    Distributions of band-energy MAE for PBE, Geometry GNN, and \ourmodel{} in the main held-out test cohort of 693 entries and a separate benchmark of six literature-validated strongly correlated materials \citep{Anisimov1991Mott, Himmetoglu2014LDAU, Tsujioka1997CrO2, Korotin1998CrO2, Saitoh1997LaCoO3, Haverkort2006LaCoO3, Ouellette2010LaNiO3, Karner2019LaNiO3, Roth1958Monoxides, Klotz2020MnO, Liu2020Monoxides, Kunes2007NiO, Morikawa1996YTiO3, Iga2004YTiO3}. 
    The correlated-material comparison is evaluated on dense Brillouin-zone meshes within the 5~eV window.
    Horizontal markers denote means.
    The annotated 9.2\% and 30.8\% reductions compare \ourmodel{} with Geometry GNN within the respective cohorts.
    \textbf{c,}
    Reconstruction errors relative to HSE06 for transition-metal $3d$ and oxygen $2p$ projected densities of states (PDOS) across six correlated materials, with spectra reconstructed using fixed PBE Mulliken projection weights to isolate differences in state-energy placement.
    Each row corresponds to one material, with PBE, Geometry GNN, and \ourmodel{} evaluated separately.
    The schematic summarizes the relative energy alignment of metal-centered transition-metal $3d$ and ligand-derived oxygen $2p$ states, which is particularly relevant to correlation-sensitive electronic structure.
    The MnO spectra in Figure~\ref{fig:overview} illustrate the corresponding orbital-dependent energy redistribution.
    }
    \label{fig:correlated}
\end{figure}

A central question is whether \ourmodel{} learns material- and state-specific changes in electronic structure rather than primarily reproducing the systematic tendency of HSE06 to increase band separations relative to PBE.
We examine this through electronic-character recovery, a separately held-out benchmark of six strongly correlated materials, and orbital-resolved reconstruction of transition-metal $3d$ and oxygen $2p$ states.

\paragraph{Recovery of electronic character}

A qualitative discrepancy occurs when PBE and HSE06 disagree on whether a material is metallic or insulating. Because HSE06 often increases the separation between occupied and unoccupied states relative to PBE, improved agreement could arise from a generic tendency to open gaps rather than from a material-specific correction. Thus, we classify the test entries according to their PBE-to-HSE06 electronic-character change as insulator-to-insulator, metal-to-insulator, and metal-to-metal systems (Fig.~\ref{fig:correlated}a). Across the three trained models, \ourmodel{} recovers the corresponding HSE06 electronic character with accuracies of $99.8\pm0.3\%$, $86.2\pm6.0\%$, and $79.6\pm1.7\%$, respectively. The ability to recover metal-to-insulator cases while preserving metal-to-metal systems indicates that the learned correction is material dependent rather than a uniform tendency to open band gaps.

\paragraph{Strongly correlated material evaluation}

A separate set of six strongly correlated materials was excluded from training and reserved a priori for evaluation because these correlation-sensitive systems are well established as challenging for semilocal density-functional approximations \citep{Anisimov1991Mott, Himmetoglu2014LDAU, Tsujioka1997CrO2, Korotin1998CrO2, Saitoh1997LaCoO3, Haverkort2006LaCoO3, Ouellette2010LaNiO3, Karner2019LaNiO3, Roth1958Monoxides, Klotz2020MnO, Liu2020Monoxides, Kunes2007NiO, Morikawa1996YTiO3, Iga2004YTiO3}. For each material, the represented magnetic and electronic state was additionally checked against the corresponding literature-supported phase (Methods~\ref{sec:methods_downstream}; Fig.~\ref{fig:correlated}b).

The six strongly correlated materials are evaluated on dense Brillouin-zone meshes using the representative model.
Within the 5~eV window, the material-balanced band-energy MAE decreases from 1.532~eV for PBE to 0.479~eV with \ourmodel{}, corresponding to a 68.7\% reduction.
The matched Geometry GNN reaches 0.692~eV, giving \ourmodel{} a further 30.8\% reduction in error and indicating that explicit orbital information provides additional predictive value in this correlation-sensitive regime.

The improvement also extends to the orbital-resolved electronic structure (Fig.~\ref{fig:correlated}c).
Within the $\pm5$~eV PDOS window, the material-balanced mean normalized $L_1$ error for transition-metal $3d$ states decreases from 1.451 with PBE to 0.709 with \ourmodel{}, while Geometry GNN reaches 0.989.
The corresponding oxygen $2p$ error decreases from 1.036 for PBE to 0.631 for \ourmodel{}, compared with 0.753 for Geometry GNN. These values correspond to reductions relative to PBE of 51.1\% and 39.1\% for the $3d$ and $2p$ channels, respectively, with further reductions of 28.3\% and 16.2\% relative to the geometry-based model. \ourmodel{} outperforms Geometry GNN for the transition-metal $3d$ channel in all six materials, whereas the oxygen $2p$ improvement is more material dependent.
Because the model PDOS is reconstructed using fixed PBE Mulliken projection weights transported to the predicted state energies, these results quantify improvements in orbital-resolved state-energy placement rather than prediction of target-theory wavefunctions or changes in hybridization.

\paragraph{Orbital-dependent electronic structure of MnO}

For a closer examination, Figure ~\ref{fig:overview}c shows antiferromagnetic rocksalt MnO with AFM-II ordering along $[111]$ \citep{Roth1958Monoxides,Liu2020Monoxides,Klotz2020MnO}.
The reconstructed spectrum shows distinct corrections across the Mn $3d$ and O $2p$ states.
Occupied and unoccupied Mn $3d$ features undergo different energy shifts across the two spin channels, while the O $2p$ spectral distribution is displaced toward the corresponding HSE06 energy range.
The correction therefore modifies the relative placement of metal- and ligand-derived states rather than acting only through a uniform change in the fundamental gap.
Such $d$--$p$ energetics are particularly relevant in correlated oxides, where the relative placement of transition-metal $d$ and ligand $p$ states is central to Mott and charge-transfer-sensitive electronic structure.
These orbital-resolved corrections could therefore support screening of strongly correlated materials in which doping, pressure, or strain reorganizes transition-metal $d$ and ligand $p$ states in ways relevant to emergent superconducting phases.

\subsection{Derived electronic properties}

\begin{figure}[ht!]
    \centering
    \includegraphics[width=\linewidth]{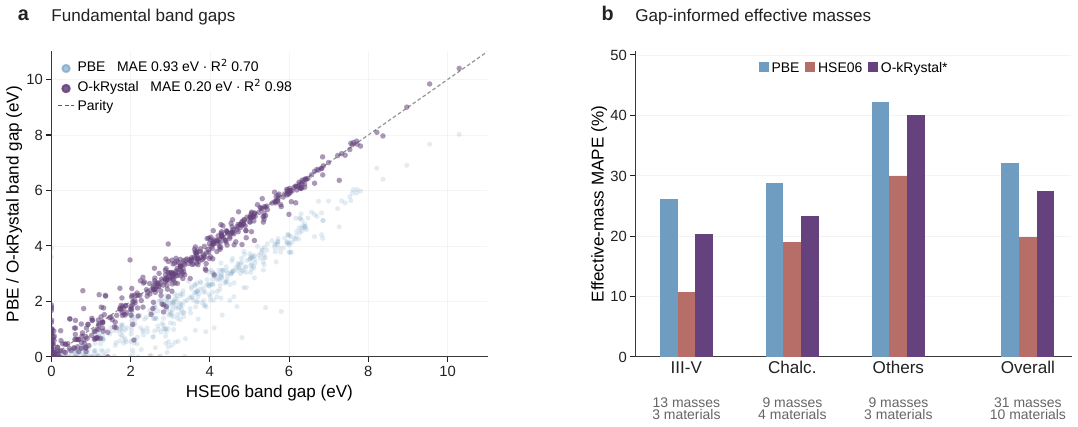}
    \caption{\textbf{Band gaps and gap-informed effective-mass estimates.}
    \textbf{a,}
    Fundamental band gaps derived from PBE and \ourmodel{} band energies compared with HSE06 reference gaps across the test set. 
    The dashed line denotes parity, and the corresponding MAE and $R^2$ values are reported for both methods.
    \textbf{b,}
    Material-balanced mean absolute percentage errors (MAPE) relative to experiment for 31 directional effective masses across ten bulk materials from mstar60 benchmark \citep{laurien2022mstar60_dataset}.
    Results are grouped into III--V semiconductors, chalcogenides, and other semiconductors.
    \ourmodel{}* denotes a calibrated gap-informed reconstruction in which the corresponding PBE effective mass for the same carrier branch, band extremum, and crystallographic direction provides the baseline curvature information and is rescaled using the predicted-to-PBE fundamental-gap ratio.
    Reconstruction parameters are fitted to HSE06 masses from the other material families in a leave-one-family-out protocol.
    }
    \label{fig:properties}
\end{figure}

The predicted state energies also enable evaluation of derived electronic properties. Fundamental band gaps are obtained directly from the predicted valence- and conduction-band edges, while carrier effective masses are estimated using a gap-dependent reconstruction motivated by the Kane model.

\paragraph{Fundamental band gaps}

Across the test set, \ourmodel{} systematically improves the fundamental band gaps relative to PBE (Fig.~\ref{fig:properties}a). The PBE gaps exhibit an MAE of 0.926~eV relative to HSE06 and an $R^2$ of 0.70. In comparison, gaps derived from the \ourmodel{}-predicted band energies reduce the MAE to 0.203~eV and increase $R^2$ to 0.98. The band gap is recovered directly from the predicted valence- and conduction-band energies rather than from a separate scalar output head. Its improved accuracy therefore shows that the state-resolved predictions remain consistent at the band edges.

\paragraph{Comparison with experimental band gaps}

We next compare with experimental measurements using the mstar60 benchmark \citep{laurien2022mstar60_dataset}, using the ten-material bulk cohort defined in Methods~\ref{sec:methods_downstream} (Figs.~\ref{fig:overview}d and \ref{fig:properties}b). PBE generally underestimates the experimental gaps, whereas HSE06 provides closer agreement. \ourmodel{} follows these material-dependent HSE06 corrections, improving agreement with experiment where HSE06 does while retaining similar deviations where HSE06 itself remains inaccurate.

\paragraph{Gap-informed effective masses}

Because the available HSE06 data do not densely resolve the local band curvature around the extrema, \ourmodel{} is not evaluated as a direct effective-mass predictor. Instead, its predicted HSE06 gaps are propagated through a gap-dependent reconstruction motivated by the Kane model (Methods~\ref{sec:methods_downstream}) to estimate 31 directional effective masses across the ten-material benchmark (Fig.~\ref{fig:properties}b). Using leave-one-family-out calibration, the reconstructed masses reduce the overall experimental MAPE from 32.05\% for PBE to 27.05\%, compared with 19.87\% for HSE06. The improvement is consistent across all three material families, indicating that more accurate band-edge energetics can improve downstream effective-mass estimates even when the local curvature itself is not directly supervised.

\subsection{Extension to \gzwz{} quasiparticle energies}

\begin{figure}[ht!]
    \centering
    \includegraphics[width=\linewidth]{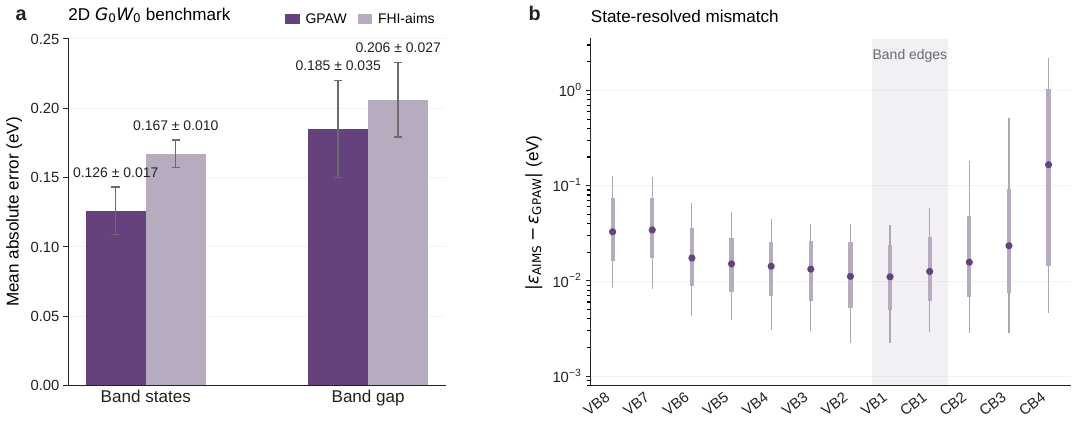}
    \caption{\textbf{Extension to \gzwz{} quasiparticle energies.}
    \textbf{a,}
    Five-fold cross-validated quasiparticle-state and band-gap MAEs for \ourmodel{} on 286 two-dimensional semiconductors from the C2DB-derived \gzwz{} benchmark \citep{Haastrup2018C2DB, gjerding2021c2db, knosgaard2022gw_ml}.
    Separately trained models use either GPAW or independently calculated FHI-aims PBE eigenvalues \citep{Mortensen2024GPAW, Blum2009FHIaims}, while the \gzwz{} labels, sampled $\ks$ points, and orbital-matrix inputs are held fixed.
    Each outer split uses three folds for training, one for validation and one for testing.
    Bars and error bars denote means and sample standard deviations across the five test folds.
    \textbf{b,}
    Absolute differences between FHI-aims and GPAW PBE eigenvalues for the eight highest valence (VB1--VB8) and four lowest conduction (CB1--CB4) states.
    Circles indicate the median discrepancy, boxes the interquartile range, and whiskers the 10th--90th percentile range.
    Shading highlights the frontier bands VB1 and CB1.
    }
    \label{fig:gw_correction}
\end{figure}

The preceding results focus on PBE-to-HSE06 corrections in three-dimensional periodic materials. We finally test whether the same orbital-learning framework extends to a physically distinct electronic-structure setting by considering \gzwz{} quasiparticle energies. Whereas HSE06 remains a generalized Kohn-Sham description, \gzwz{} introduces an energy-dependent electronic self-energy to approximate electron addition and removal energies, providing a more direct connection to photoemission observables \citep{Hedin1965GW, golze2019gw_review}. We evaluate \ourmodel{} on a C2DB-derived quasiparticle benchmark comprising 286 two-dimensional semiconductors using the cross-validation protocol described in Methods~\ref{sec:methods_gw} \citep{Haastrup2018C2DB,gjerding2021c2db,knosgaard2022gw_ml}.

\paragraph{Beyond DFT band structure prediction}

Using five-fold material-level cross-validation with a separate validation fold for checkpoint selection, \ourmodel{} achieves a quasiparticle-state MAE of $0.126\pm0.017$~eV across the twelve available bands and a band-gap MAE of $0.185\pm0.035$~eV when using the GPAW PBE baseline (Fig.~\ref{fig:gw_correction}a; Supplementary Table~\ref{tab:s7}). These results extend the same orbital-learning formulation from three-dimensional PBE-to-HSE06 band reconstruction to two-dimensional \gzwz{} quasiparticle energies without modifying the underlying learning architecture.

The established C2DB \gzwz{} benchmark also enables comparison with the state-specific approach of Kn{\o}sgaard and Thygesen, which used electronic state fingerprints constructed from orbital projections and operator matrix elements together with gradient-boosted regression \citep{knosgaard2022gw_ml}. Their model achieved a band-gap MAE of 0.18~eV on the selected test split and state-energy MAE of $0.13\pm0.02$~eV with repeated random splits. \ourmodel{} achieves comparable accuracy under a separate train/validation/test fold protocol, while retaining a unified orbital-learning architecture across HSE06 and \gzwz{} targets.

\paragraph{Baseline consistency in $\Delta$-learning}

Because the reference \gzwz{} calculations were performed on top of GPAW PBE electronic structures, we further examine how consistency between the baseline and target calculations affects the learned quasiparticle correction (Fig.~\ref{fig:gw_correction}a). Using the corresponding GPAW PBE baseline, \ourmodel{} achieves an all-band MAE of $0.126\pm0.017$~eV, whereas replacing it with independently calculated FHI-aims PBE bands increases the error to $0.168\pm0.010$~eV. The band-gap MAE also increases, from $0.185\pm0.035$~eV with the GPAW baseline to $0.206\pm0.027$~eV with the FHI-aims baseline \citep{Mortensen2024GPAW, Blum2009FHIaims}.

The state-resolved comparison helps explain this difference (Fig.~\ref{fig:gw_correction}b). The two PBE calculations agree most closely for the band-edge energies surrounding the band gap, whereas their discrepancy increases for states farther from the band edges, particularly for the highest-energy conduction bands. This pattern is consistent with the larger sensitivity of full state-resolved quasiparticle prediction to baseline inconsistency, while the band-gap error reflects only the frontier states. Differences between the two PBE calculations may arise from several implementation-specific factors, including the basis representation and treatment of the ionic cores. More generally, these results show that state-resolved $\Delta$-learning benefits from consistency between the baseline electronic structure and the starting point used to construct the corresponding target-theory quasiparticle calculation.

\section{Discussion}

\ourmodel{} uses lower-cost electronic structure as both the baseline for $\Delta$-learning and a representation of the underlying electronic state.
While geometry and PBE band information capture much of the systematic PBE-to-HSE06 correction, orbital features provide additional information for material- and state-specific deviations.
This advantage becomes more pronounced in the separately held-out correlated-material benchmark and is particularly consistent for transition-metal $3d$ states, suggesting that explicit electronic information is especially valuable when localized, correlation-sensitive states dominate the low-energy spectrum.
Representation comparisons likewise show that the full dual-space model achieves progressively lower errors than the geometry-based and real-space alternatives as the training set and evaluated energy range increase (Supplementary Fig.~\ref{fig:s1}).

The reciprocal-space results further show that the learned correction generalizes beyond the high-symmetry $\ks$-points used for supervision.
Accuracy remains essentially unchanged at unseen path-interior points, while dense Brillouin-zone evaluation enables reconstruction of target-level spectral quantities away from conventional band paths.
This combination of state-resolved prediction and reciprocal-space generalization allows \ourmodel{} to recover material-specific changes in electronic dispersion without requiring target-theory calculations at each queried $\ks$-point.

These capabilities provide a natural route for electronic-structure screening.
\ourmodel{} recovers both metal-to-insulator changes and systems that remain metallic, while the correlated-material analysis resolves orbital-dependent changes in the relative placement of transition-metal $d$ and ligand $p$ states.
Such predictions could identify materials for which DFT provides an inadequate low-energy electronic structure and prioritize candidates for more computationally demanding hybrid-functional or many-body calculations, including systems in which doping, pressure, or strain reorganizes metal--ligand states in ways relevant to superconducting phases and Cooper-pair formation.
Extension to \gzwz{} quasiparticle energies further shows that the state-resolved learning framework is not specific to HSE06, while the sensitivity to the PBE starting point emphasizes that the baseline electronic states should remain consistent with those from which the target correction is defined.

Several limitations define the present scope. \ourmodel{} still requires a converged baseline electronic-structure calculation and therefore occupies a different cost-accuracy regime from structure-only surrogate models.
This dependence could be substantially reduced by emerging approaches that learn the potential-to-density map and avoid repeated orbital diagonalizations during self-consistency \citep{Khan2026KSMap}.
Coupling such accelerated baseline solvers with orbital learning target-theory correction could extend the present framework to larger systems while retaining a physically grounded electronic starting point.
\ourmodel{} also predicts target-theory state energies rather than wavefunctions, so orbital-projected reconstructions retain projection information from the baseline calculation and cannot capture target-induced changes in hybridization or wavefunction character.
A natural next step is therefore to move from state-energy correction toward learning richer electronic objects, including electron densities \citep{lou2026moire}, compact target-theory Hamiltonians in Wannier representations \citep{Qi2025WANDER, tyner2026GWaNN} and, ultimately, many-body Green's functions or self-energies that provide access to quasiparticle spectra, spectral functions, and broader ground- and excited-state observables \citep{Venturella2025GreenFunction}.
In conclusion, these directions suggest a path from accelerating the baseline quantum calculation to learning increasingly complete representations of the target electronic theory.

\section{Methods}

\subsection{Dataset construction and electronic-structure data}
\label{sec:methods_data}

\paragraph{Three-dimensional PBE--HSE06 dataset}

The primary dataset is derived from paired PBE and HSE06 electronic-structure calculations for chemically diverse three-dimensional crystalline materials \citep{nair2025hse06_dataset}. After data cleaning and exclusion of materials reserved for downstream benchmarks, the final benchmark subset contains 6,900 band-structure entries corresponding to 5,988 parent materials. Spin-polarized materials are represented by separate spin-channel entries. Parent materials are removed when the absolute difference between the PBE and HSE06 magnetic moments exceeds $0.5~\mu_{\mathrm B}$, reducing inconsistencies between the baseline and target electronic states.

The final benchmark dataset is partitioned at the parent-material level into approximately 8:1:1 training, validation, and test sets, such that repeated entries and corresponding spin channels remain within the same partition. Complete dataset accounting and seed-specific split sizes are reported in Supplementary Section~\ref{sec:supp_dataset} and Supplementary Table~\ref{tab:s1}. Materials used for the correlated-material and experimental downstream benchmarks are excluded before construction of these splits.

\paragraph{State alignment and learning targets}

PBE and HSE06 states are paired according to their stored band ordering relative to the band edges. The common valence and conduction manifolds are retained, and each PBE and HSE06 spectrum is independently centered at the midpoint between the pathwise maximum of the highest valence band and minimum of the lowest conduction band before constructing the state-resolved correction

\begin{equation}
    \Delta\varepsilon_{n\mathbf{k}\sigma}
    =
    \varepsilon^{\mathrm{HSE06}}_{n\mathbf{k}\sigma}
    -
    \varepsilon^{\mathrm{PBE}}_{n\mathbf{k}\sigma}.
\end{equation}

Throughout the band-energy analysis, $\eref$ denotes this operational band-edge-energy reference rather than a separately determined thermodynamic chemical potential.

For the primary three-dimensional task, supervision is restricted to the stored high-symmetry $\ks$-points. Interior points along the corresponding paths are reserved for evaluation of reciprocal-space generalization. Band-energy errors are evaluated within PBE-side energy windows of 2, 5, 10, and 20~eV around the \ereftext{}.

\paragraph{Electronic-structure data}

The baseline orbital quantities are obtained from atom-centered electronic-structure calculations in the FHI-aims  representation \citep{Blum2009FHIaims} and include the Hamiltonian, overlap, density, and spin-density matrices described below. Orbital preprocessing and the final input representation are summarized in Supplementary Section~\ref{sec:supp_model}.

\subsection{Orbital representation}
\label{sec:methods_representation}

\paragraph{Orbital-resolved input features}

For each baseline PBE calculation, \ourmodel{} uses orbital-resolved quantities derived from an atom-centered basis representation. The input operator set comprises the Hamiltonian $\mathbf H$, overlap $\mathbf S$, density matrix $\mathbf P$, and spin-density matrix $\mathbf P_{\mathrm{spin}}$. These quantities are organized into atom-centered on-site and pairwise blocks and provide a representation of the converged baseline electronic state beyond the PBE eigenvalues alone.

For an atom-centered nonorthogonal basis, the generalized eigenvalue problem at each crystal momentum $\ks$ is written as

\begin{equation}
    \mathbf H(\mathbf k)\mathbf C(\mathbf k)
    =
    \mathbf S(\mathbf k)\mathbf C(\mathbf k)\boldsymbol{\varepsilon}(\mathbf k),
\end{equation}

where $\mathbf C(\mathbf k)$ contains the generalized eigenvectors and $\boldsymbol{\varepsilon}(\mathbf k)$ the corresponding PBE eigenvalues. The density matrix is constructed from the occupied eigenstates and subsequently represented in the atomic-orbital basis. Additional orbital-representation and preprocessing details are provided in Supplementary Section~\ref{sec:supp_model}.

\paragraph{Real-space representation}

For atomic orbitals $\mu$ and $\nu$ centered on atoms $i$ and $j$, respectively, the real-space matrix block associated with lattice translation $\mathbf R$ is

\begin{equation}
    O_{ij}^{\mu\nu}(\mathbf R)
    =
    \left\langle
    \phi_i^\mu(\mathbf 0)
    \middle|
    \hat O
    \middle|
    \phi_j^\nu(\mathbf R)
    \right\rangle,
\end{equation}

where $\hat O\in\{\hat H,\hat S,\hat P,\hat P_{\mathrm{spin}}\}$. On-site blocks are retained separately, while pairwise blocks encode interactions between orbitals on neighboring periodic images. The neural representation uses atom-centered orbital blocks padded to a maximum dimension of 16 per atom. The real-space representation is organized as a periodic directed atom-image graph. Each retained pair edge carries the corresponding orbital matrix blocks, while on-site blocks are presented separately. The preprocessed pair lists contain up to 32 electronically selected pair blocks per central atom. The reciprocal-space blocks are associated with the same retained atom-image pairs as the real-space representation, with separate $\ks$-dependent on-site blocks.

\paragraph{Reciprocal-space representation}

The corresponding reciprocal-space representation is related to the real-space matrix elements through the Bloch transform,
\begin{equation}
O_{ij}^{\mu\nu}(\ks)
=
\sum_{\rs}
e^{i\ks\cdot\rs}
O_{ij}^{\mu\nu}(\rs).
\end{equation}
In the final model, the real components of $\mathbf H(\ks)$, $\mathbf S(\ks)$, $\mathbf P(\ks)$, and $\mathbf P_{\mathrm{spin}}(\ks)$ are used as reciprocal-space orbital features.

\paragraph{Bloch-phase conditioning}

$\ks$ dependence is additionally introduced on each periodic pair through the full atom-to-image displacement,

\begin{equation}
    \mathbf r_{ij\mathbf R}
    =
    \mathbf r_j-\mathbf r_i+\mathbf R,
\end{equation}

using

\begin{equation}
    \left[
    \sin\!\left(\mathbf k\cdot\mathbf r_{ij\mathbf R}\right),
    \cos\!\left(\mathbf k\cdot\mathbf r_{ij\mathbf R}\right)
    \right].
\end{equation}

These phase factors condition the local representation on the queried crystal momentum. The full \ourmodel{} representation combines the real-space orbital blocks, Bloch-phase conditioning, and explicit reciprocal-space orbital features. A corresponding real-space + phase ablation removes the explicit reciprocal-space matrices retaining the remaining inputs unchanged.

\subsection{\ourmodel{} architecture}
\label{sec:methods_architecture}

\paragraph{Periodic equivariant message passing}

\ourmodel{} is built on a periodic $\mathrm{E}(3)$-equivariant graph neural network that processes the orbital blocks associated with atoms and periodic atom pairs. On-site orbital blocks initialize atom-centered features, while pairwise blocks contribute to equivariant message passing through periodic relative displacements and spherical-harmonic geometric features. Atomic species are incorporated through learned element embeddings, and the resulting atom-wise representations are pooled into a material- and $\ks$-dependent latent representation.

The final model comprises two jointly trained branches: a real-space + Bloch-phase affine branch and a dual-space residual branch that additionally receives explicit reciprocal-space orbital features. The affine branch inherits four equivariant interaction layers and four decoding layers from the model defaults, whereas the residual branch explicitly uses two interaction and two decoding layers. Real-space orbital features and Bloch-phase information enter the affine branch, while explicit reciprocal-space orbital features provide an additional $\ks$-resolved residual contribution. Architectural and training hyperparameters are summarized in Supplementary Section~\ref{sec:supp_model} and Supplementary Table~\ref{tab:s2}

For the representation ablation, we additionally train a Real+Bloch-phase variant that retains the real-space orbital blocks and Bloch-phase conditioning while omitting the explicit reciprocal-space operator matrices. All other training and evaluation settings are matched to the full model (Supplementary Fig.~\ref{fig:s1}).

\paragraph{Band-query representation}

Rather than predicting a fixed-dimensional band vector, \ourmodel{} represents each electronic state through a signed band query defined relative to the band edges. Valence states are assigned negative indices and conduction states positive indices, allowing a variable number of valence and conduction bands to be queried across materials.

Each band query is combined with five deterministic descriptors derived from the corresponding PBE band over the retained high-symmetry-point set: the mean energy, standard deviation, minimum energy, maximum energy, and band width. Using the PBE band-edge reference

\[
\eref^{\mathrm{PBE}}
=
\frac{1}{2}
\left[
\max_{\ks}\varepsilon^{\mathrm{PBE}}_{\mathrm{VB1},\ks}
+
\min_{\ks}\varepsilon^{\mathrm{PBE}}_{\mathrm{CB1},\ks}
\right],
\]

the descriptor vector for band \(n\) is

\[
\mathbf{f}_n =
\left[
\frac{\mu_n-\eref^{\mathrm{PBE}}}{10},
\frac{\sigma_n}{5},
\frac{\varepsilon_{n,\min}^{\mathrm{PBE}}-\eref^{\mathrm{PBE}}}{10},
\frac{\varepsilon_{n,\max}^{\mathrm{PBE}}-\eref^{\mathrm{PBE}}}{10},
\frac{\varepsilon_{n,\max}^{\mathrm{PBE}}-\varepsilon_{n,\min}^{\mathrm{PBE}}}{5}
\right],
\]

with all input energies in eV. These descriptors are supplied to the affine decoder with the signed-band embedding.

The descriptors are computed over valid high-symmetry points when available. If no retained high-symmetry point is present for a sample, all valid retained $\ks$-points are used.

\paragraph{Affine band correction and dual-space residual}

The state-resolved correction is decomposed into an affine band contribution and a $\ks$-dependent residual. For band $n$, the affine contribution is written as

\begin{equation}
    \widehat{\Delta\varepsilon}^{\mathrm{aff}}_{n\mathbf{k}}
    =
    c_n
    +
    b_n
    \left(
    \varepsilon^{\mathrm{PBE}}_{n\mathbf{k}}
    -
    \overline{\varepsilon}^{\mathrm{PBE}}_n
    \right),
\end{equation}

where $c_n$ and $b_n$ are predicted from the pooled material representation and band-query features, and $\overline{\varepsilon}^{\mathrm{PBE}}_n$ denotes the mean PBE band energy over the available high-symmetry points. This term captures band-dependent shifts and changes in dispersion relative to PBE.

A second branch predicts a pointwise residual $r_{n\mathbf{k}}$ from the combined real- and reciprocal-space orbital representation. To avoid redundancy with the affine contribution, the residual is projected at the supervised high-symmetry points to remove its mean and its component linear in the PBE dispersion, yielding 
\begin{equation}
    r^{\perp}_{n\mathbf{k}}
    =
    r_{n\mathbf{k}}
    -
    \overline{r}_{n}
    -
    a_n
    \left(
    \varepsilon^{\mathrm{PBE}}_{n\mathbf{k}}
    -
    \overline{\varepsilon}^{\mathrm{PBE}}_{n}
    \right),
\end{equation}

where $\overline{r}_{n}$ is the mean residual over the supervised high-symmetry points and $a_n$ is the least-squares coefficient of the residual along the centered PBE dispersion. The final correction is then

\begin{equation}
    \widehat{\Delta\varepsilon}_{n\mathbf{k}}
    =
    \widehat{\Delta\varepsilon}^{\mathrm{aff}}_{n\mathbf{k}}
    +
    r^{\perp}_{n\mathbf{k}}.
\end{equation}

By construction, the affine branch captures band-dependent shifts and changes correlated with the PBE dispersion, while the orthogonalized residual branch represents the remaining $\ks$-dependent correction beyond these components.

The target-theory energy is reconstructed as

\begin{equation}
    \widehat{\varepsilon}^{\mathrm{target}}_{n\mathbf{k}}
    =
    \varepsilon^{\mathrm{PBE}}_{n\mathbf{k}}
    +
    \widehat{\Delta\varepsilon}_{n\mathbf{k}}.
\end{equation}

Both branches are optimized jointly. The model predicts state energies only and does not predict target-theory wavefunctions or orbital-projection coefficients.

\subsection{Learning objective and optimization}

\paragraph{State-resolved $\Delta$-learning}

\ourmodel{} is trained to predict the state-resolved correction between the baseline and target electronic-structure methods,

\begin{equation}
    \Delta\varepsilon_{n\mathbf{k}\sigma}
    =
    \varepsilon^{\mathrm{target}}_{n\mathbf{k}\sigma}
    -
    \varepsilon^{\mathrm{PBE}}_{n\mathbf{k}\sigma}.
\end{equation}

For the primary three-dimensional task, the target theory is HSE06. PBE and HSE06 states are paired by their stored band ordering relative to the band edges, and unmatched or padded states are excluded from the loss.

\paragraph{Training objective}

Let $m$, $n$, $\mathbf{k}$, and $\sigma$ index the band-structure entry, band, crystal momentum, and spin channel, respectively. For each valid PBE--target state pair, we define the prediction residual as
$e_{mn\mathbf{k}\sigma}
=
\widehat{\Delta\varepsilon}_{mn\mathbf{k}\sigma}
-
\Delta\varepsilon_{mn\mathbf{k}\sigma}$
and penalize it using a SmoothL1 loss with transition parameter
$\kappa=0.1$~eV.

To emphasize states near the PBE reference energy while retaining supervision over the broader electronic spectrum, each valid state is assigned the weight
\begin{equation}
w_{mn\mathbf{k}\sigma}
=
\exp\left[
-\left(
\frac{
\varepsilon^{\mathrm{PBE}}_{mn\mathbf{k}\sigma}
-
\eref^{\mathrm{PBE}}
}{
E_w
}
\right)^2
\right],
\qquad
E_w=10~\mathrm{eV}.
\end{equation}
The state-energy loss is the normalized weighted SmoothL1 error,
\begin{equation}
\mathcal{L}_{\mathrm{state}}
=
\frac{
\displaystyle
\sum_{(m,n,\mathbf{k},\sigma)\in\mathcal{V}}
w_{mn\mathbf{k}\sigma}\,
\rho_{\kappa}\!\left(e_{mn\mathbf{k}\sigma}\right)
}{
\displaystyle
\sum_{(m,n,\mathbf{k},\sigma)\in\mathcal{V}}
w_{mn\mathbf{k}\sigma}
},
\end{equation}
where $\mathcal{V}$ denotes the set of valid matched states in the training batch and $\rho_{\kappa}$ denotes the SmoothL1 penalty.

To penalize systematic bandwise shifts, we additionally apply an offset loss to the mean signed residual of each band over the supervised high-symmetry
points,
\begin{equation}
\mathcal{L}_{\mathrm{offset}}
=
\frac{1}{|\mathcal{B}|}
\sum_{(m,n,\sigma)\in\mathcal{B}}
\rho_{\kappa}\!\left(
\frac{1}{N^{\mathrm{HS}}_{mn\sigma}}
\sum_{\mathbf{k}\in\mathrm{HS}}
e_{mn\mathbf{k}\sigma}
\right),
\end{equation}
where $\mathcal{B}$ is the set of valid entry--band--spin combinations and $N^{\mathrm{HS}}_{mn\sigma}$ is the number of supervised high-symmetry points for each combination.

An indirect-gap loss additionally penalizes errors in the reconstructed fundamental gap,
\begin{equation}
\mathcal{L}_{\mathrm{gap}}
=
\frac{1}{|\mathcal{G}|}
\sum_{m\in\mathcal{G}}
\rho_{\kappa}\!\left(
\widehat{E}_{g,m}
-
E^{\mathrm{target}}_{g,m}
\right),
\end{equation}
where $\widehat{E}_{g,m}$ and $E^{\mathrm{target}}_{g,m}$ are the predicted and reference fundamental gaps obtained from the corresponding valence- and conduction-band extrema, and $\mathcal{G}$ contains entries for which both extrema are valid.

The final training objective is
\begin{equation}
\mathcal{L}
=
\mathcal{L}_{\mathrm{state}}
+
\mathcal{L}_{\mathrm{offset}}
+
0.2\mathcal{L}_{\mathrm{gap}}.
\end{equation}
All other auxiliary objectives, including direct-gap, dispersion, curvature, ranking, and effective-mass terms, have zero weight in the production model. The state-energy correction targets are expressed directly in eV without standardization. Additional training settings are summarized in Supplementary Section~\ref{sec:supp_model} and Supplementary Table~\ref{tab:s2}.

\paragraph{Optimization and checkpoint selection}

Models are optimized with AdamW using an initial learning rate of $5\times10^{-4}$, weight decay of 0.005, and momentum parameters $(\beta_1,\beta_2)=(0.9,0.95)$. Training is performed for 100 epochs with a 20-epoch linear warm-up followed by cosine learning-rate decay to $10^{-5}$. Training uses dynamic batching with between one and twelve entries per batch, subject to a maximum of 700 atoms. Additional pair- and $\ks$-pair padding limits are imposed to control memory usage during training.

Three independently trained models are used for the principal three-dimensional results, with seeds 42, 43, and 44. Within each seed, the learning-curve subsets are nested while the validation and test populations remain fixed across training-set sizes. The model checkpoint used for evaluation is selected by the lowest unweighted mean validation MAE across the 2, 5, 10, and 20~eV energy windows.

\subsection{Baseline models}
\label{sec:methods_baseline}

\paragraph{Scissor and Linear+KRR baselines}

We compare \ourmodel{} against progressively more expressive corrections to the PBE band energies. The scissor baseline applies separate global mean corrections to valence- and conduction-side states, estimated from the training set.  The Linear+KRR baseline uses six PBE-derived state and material descriptors in a linear model followed by RBF kernel-ridge regression of the remaining residual. For the primary comparison, both baselines are fitted using only the high-symmetry-point states of the corresponding training partition and evaluated on the matched high-symmetry-point test subset using the same PBE-side reference and energy-window conventions as \ourmodel{}. Implementation details are provided in Supplementary Section~\ref{sec:supp_model} and Supplementary Table~\ref{tab:s3}.

\paragraph{Geometry GNN baseline}

To compare against a geometry-based correction supplied with the PBE band baseline, we train a graph neural network using the same seed-specific data splits and state-resolved prediction targets. The model receives atomic species, periodic pair distances, Bloch-phase features, and the PBE band energies and descriptors used by the band decoder, but no orbital matrices. Unlike the electronically preselected pair graph used by the orbital models, the Geometry GNN constructs its periodic graph directly from atomic positions, retaining up to 32 nearest atom-image neighbors per central atom by distance. Its periodic message-passing network therefore learns the PBE-to-HSE06 correction from geometric information together with the baseline band structure. The Geometry GNN inputs and their distinction from the orbital model are summarized in Supplementary Section~\ref{sec:supp_model} and Supplementary Table~\ref{tab:s3}.

\paragraph{Real-space + Bloch-phase ablation}

To quantify the contribution of explicit reciprocal-space orbital information, we compare the full dual-space \ourmodel{} representation with a Real+Bloch-phase variant that retains the real-space orbital blocks, atomic structure, PBE band information, band-query representation, and Bloch-phase conditioning while omitting the explicit reciprocal-space orbital matrices. Both representations are trained using the same optimization settings, loss function, checkpoint-selection criterion, seed-specific data splits, and training-set sizes. The comparison is evaluated across approximately 0.5k, 2.0k, 3.5k, and 5.5k training entries for seeds 42, 43, and 44. Results are reported in Supplementary Section~\ref{sec:supp_reciprocal_representation} and Supplementary Figure~\ref{fig:s1}.

\subsection{Evaluation}
\label{sec:methods_evaluation}

\paragraph{Band-energy errors}

State-resolved prediction errors are evaluated as the mean absolute error between predicted and target-theory band energies over valid matched states. For a band-structure entry $m$ and energy window $W$

\begin{equation}
    \mathrm{MAE}_m(W)
    =
    \frac{1}{|V_m(W)|}
    \sum_{(n,\mathbf{k})\in V_m(W)}
    \left|
    \widehat{\varepsilon}_{mn\mathbf{k}}
    -
    \varepsilon^{\mathrm{target}}_{mn\mathbf{k}}
    \right|,
\end{equation}

where $V_m(W)$ denotes the set of valid states satisfying the corresponding PBE-side energy-window criterion
\[
|E^{\mathrm{PBE}}-\eref^{\mathrm{PBE}}|< W.
\]
Entries for which $V_m(W)$ is empty are excluded from the corresponding window average for all methods. Unless otherwise stated, reported three-dimensional band-energy errors are averaged over band-structure entries, with spin channels treated as separate entries. For the main HSE06 benchmark, $W\in\{2,5,10,20\}$~eV.

\paragraph{Evaluation checkpoint}

The primary benchmark in Fig.~\ref{fig:results} reports results across three separately trained models using distinct parent-level data splits. Unless otherwise stated, the detailed reciprocal-space and downstream analyses in Figs.~\ref{fig:generalization}, \ref{fig:correlated}, and \ref{fig:properties} use the provided seed-42 \texttt{best\_avg\_win} checkpoint.

\paragraph{Reciprocal-space generalization}

For the three-dimensional HSE06 task, \ourmodel{} is supervised only at the stored high-symmetry $\ks$-points. Reciprocal-space generalization is evaluated on fixed query sets containing previously unseen path-interior points, using the same state-matching and PBE-side energy-window conventions as the primary band-energy evaluation. Individual predictions are therefore conditioned on the supplied $\ks$-point set rather than interpreted as query-set-independent single-point predictions. Path-interior diagnostics, including the difference between interior-point and high-symmetry-point errors and the dependence on reciprocal-space distance, relative segment position, and segment length, are reported in Supplementary Section~\ref{sec:supp_k_generalization} and Supplementary Fig.~\ref{fig:s2}. Representative full-path predictions are shown in Supplementary Fig.~\ref{fig:s3}.

\paragraph{Density of states reconstruction}

To assess predictions away from conventional high-symmetry paths, \ourmodel{} is evaluated on dense, material-specific regular $\ks$-point meshes throughout the Brillouin zone. Predicted state energies are used to reconstruct the density of states using tetrahedron integration followed by Gaussian broadening with $\sigma=0.10$~eV on a common energy grid from $-10$ to $10$~eV.

The DOS reconstruction error within $W=5$~eV of the \ereftext{} is quantified using the normalized $L_1$ difference

\begin{equation}
    D_m
    =
    \frac{
    \int_{-W}^{W}
    \left|
    g_m(E)-g_m^{\mathrm{HSE06}}(E)
    \right|
    \,dE
    }{
    \int_{-W}^{W}
    g_m^{\mathrm{HSE06}}(E)
    \,dE
    },
\end{equation}

where $g_m(E)$ and $g_m^{\mathrm{HSE06}}(E)$ denote the predicted and reference densities of states, respectively. Spin-resolved contributions are combined at the parent-material level before evaluating the DOS error. This metric evaluates aggregate spectral agreement over the energy window rather than pointwise correspondence of individual off-path eigenstates.

\paragraph{Electronic-character classification}

Electronic character is determined from the indirect band gap of each band-structure entry. An entry is classified as insulating when the corresponding physical gap is positive and metallic otherwise; the predefined odd-electron metallicity convention is applied to unpolarized entries. Spin channels are classified separately, while a parent material is metallic if any constituent spin channel is metallic. We categorize the baseline-to-target change as insulator-to-insulator, metal-to-insulator, metal-to-metal, or insulator-to-metal, and compare the \ourmodel{} classification with the corresponding HSE06 reference.

\subsection{Downstream benchmarks}
\label{sec:methods_downstream}

\paragraph{Correlated-material benchmark and band-energy evaluation}

A separate literature-curated benchmark is used to evaluate \ourmodel{} in materials with localized and correlation-sensitive electronic structure.
Six parent materials are retained after requiring consistency between the represented magnetic and electronic state and the corresponding literature-supported phase.
The cohort comprises the correlated ferromagnetic metal CrO$_2$ \citep{Tsujioka1997CrO2,Korotin1998CrO2}, low-spin LaCoO$_3$ \citep{Saitoh1997LaCoO3,Haverkort2006LaCoO3}, correlated metallic LaNiO$_3$ \citep{Ouellette2010LaNiO3,Karner2019LaNiO3}, AFM-II MnO and NiO \citep{Roth1958Monoxides,Liu2020Monoxides,Klotz2020MnO,Kunes2007NiO}, and ferromagnetic Mott-insulating YTiO$_3$ \citep{Morikawa1996YTiO3,Iga2004YTiO3}.
All six materials are excluded before construction of the primary training, validation, and test splits.

Band-energy errors are evaluated using the representative seed-42 model on dense, material-specific Brillouin-zone meshes.
For material $m$, the mean absolute error within an energy window $W$ is
\begin{equation}
\mathrm{MAE}_m(W)
=
\frac{1}{|\mathcal{V}_m(W)|}
\sum_{(n,\ks,\sigma)\in\mathcal{V}_m(W)}
\left|
\hat{\varepsilon}_{mn\ks\sigma}
-
\varepsilon^{\mathrm{HSE06}}_{mn\ks\sigma}
\right|,
\end{equation}
where $\mathcal{V}_m(W)$ contains the retained states satisfying
\begin{equation}
\left|
\varepsilon^{\mathrm{PBE}}_{mn\ks\sigma}
-
E^{\mathrm{PBE}}_{\mathrm{mid},m}
\right|
\leq W .
\end{equation}
The main-text comparison uses $W=5$~eV, while $W=20$~eV is used as a wider-window evaluation.
No reciprocal-space quadrature weights are applied to the band-energy MAE.
For spin-polarized materials, retained states from both spin channels are pooled before computing the material-level error.
Aggregate values are material balanced by assigning equal weight to each of the six parent materials.

\paragraph{Orbital-resolved projected density of states}

Orbital-resolved electronic structure is evaluated for the transition-metal $3d$ and oxygen $2p$ channels.
To isolate differences in state-energy placement, all projected densities of states are constructed using the same Mulliken orbital weights obtained from the baseline PBE wavefunctions.
For a set of state energies $\varepsilon_{mn\ks\sigma}^{X}$ from method $X$,
\begin{equation}
g^{X}_{\alpha,m}(E)
=
\sum_{n\ks\sigma}
w_{\ks}\,
p^{\mathrm{PBE}}_{\alpha,mn\ks\sigma}\,
\delta\!\left(
E-\varepsilon^{X}_{mn\ks\sigma}
\right),
\end{equation}
where $\alpha$ denotes the orbital channel, $p^{\mathrm{PBE}}_{\alpha,mn\ks\sigma}$ is the corresponding PBE Mulliken projection weight, and $w_{\ks}$ is the reciprocal-space integration weight.
For PBE, Geometry GNN, \ourmodel{}, and HSE06, only the state energies entering the reconstruction differ.
The PDOS is evaluated on dense reciprocal-space meshes using linear tetrahedron integration and subsequently broadened with a Gaussian of width $\sigma=0.10$~eV.
This construction therefore compares orbital-resolved energy distributions at fixed orbital character and does not evaluate target-theory changes in wavefunctions or hybridization.

Agreement with the HSE06 energy distribution is quantified using the normalized $L_1$ error
\begin{equation}
D_{\alpha,m}
=
\frac{
\displaystyle
\int_{-5\,\mathrm{eV}}^{5\,\mathrm{eV}}
\left|
g^{\mathrm{method}}_{\alpha,m}(E)
-
g^{\mathrm{HSE06}}_{\alpha,m}(E)
\right|\,dE
}{
\displaystyle
\int_{-5\,\mathrm{eV}}^{5\,\mathrm{eV}}
\left|
g^{\mathrm{HSE06}}_{\alpha,m}(E)
\right|\,dE
},
\end{equation}
where energies are referenced to $E_{\mathrm{mid}}$.
The metric is dimensionless, and aggregate values are obtained by averaging the six material-level errors with equal weight.

\paragraph{Experimental band-gap benchmark}

Experimental comparisons use the mstar60 benchmark \citep{laurien2022mstar60_dataset}. The primary bulk cohort contains ten materials for which the required PBE, HSE06, and experimental quantities are available and whose constituent elements are represented in the three-dimensional training set. Monolayer systems are excluded from the bulk comparison, Pb chalcogenides are excluded from the primary benchmark because Pb is absent from the training chemistry, and black phosphorus is excluded because the required PBE effective masses are unavailable. The benchmark composition and inclusion criteria are summarized in Supplementary Section~\ref{sec:supp_mstar60} and Supplementary Table~\ref{tab:s5}.

\paragraph{Gap-informed effective-mass reconstruction}

Effective masses are not predicted directly from \ourmodel{} band curvature. Instead, for observation $i$, the carrier effective mass is predicted through the reconstruction

\begin{equation}
\hat{m}^{*}_{i}
=
m^{*}_{i,\mathrm{PBE}}\,r_i^{\beta_i},
\end{equation}
where $\hat{m}^{*}_{i}$ is the predicted carrier effective mass, $m^{*}_{i,\mathrm{PBE}}$ is the PBE effective mass for the same carrier branch, band extremum, and crystallographic direction as observation $i$, $r_i$ is a regularized ratio between the \ourmodel{}-predicted and PBE fundamental band gaps and $\beta_i$ is a mass-dependent exponent whose parameters are calibrated without the held-out material family.

For this downstream analysis, the PBE and predicted HSE06 fundamental gaps are evaluated over the complete stored band path as the conduction-band minimum minus the valence-band maximum. The reconstruction is evaluated for 31 directional effective-mass observations across ten materials, grouped into III--V semiconductors, chalcogenides, and other semiconductors. To assess transfer across chemical families, each family is held out in turn while the reconstruction parameters are calibrated using HSE06 effective masses from the remaining families. For evaluation, the absolute percentage errors are first averaged over all retained effective-mass observations belonging to each material, and the resulting material-level errors are averaged equally across materials. Reported family and overall MAPE values are therefore material balanced. Experimental masses are excluded from calibration and are used only for final evaluation. The exact regularization, calibration procedure, family-specific parameters, benchmark composition, and all 31 effective-mass observations are provided in Supplementary Section~\ref{sec:supp_mstar60} and Supplementary Tables~\ref{tab:s5} and \ref{tab:s6}.

\subsection{$G_0W_0$ quasiparticle benchmark}
\label{sec:methods_gw}

\paragraph{Two-dimensional quasiparticle dataset}

We evaluate the same orbital learning framework on a two-dimensional \gzwz{} quasiparticle benchmark derived from C2DB \citep{Haastrup2018C2DB, gjerding2021c2db, knosgaard2022gw_ml}. The final cohort contains 286 nonmagnetic semiconductor records with quasiparticle energies for the eight highest valence states and four lowest conduction states. The target quantities are \gzwz{} quasiparticle energies constructed from GPAW PBE starting points. We compare two PBE eigenvalue baselines: the corresponding GPAW spectrum and an independently calculated FHI-aims spectrum \citep{Mortensen2024GPAW, Blum2009FHIaims}. The \gzwz{} labels, $\ks$-point sets, and real- and reciprocal-space orbital inputs are held fixed between the two variants. Changing the PBE eigenvalues consequently changes the $\Delta$-learning targets. Separate models are trained from scratch for the two baselines using otherwise matched settings.

The \gzwz{} task is trained separately from the three-dimensional HSE06 model while retaining the same underlying orbital learning architecture and state-resolved $\Delta$-learning formulation. Unlike the primary three-dimensional task, the \gzwz{} training queries include the available high-symmetry points together with up to nine evenly spaced stored interior points per path segment. Five-fold cross-validation is performed with separate training, validation, and test folds. For each outer fold, three source folds are used for training, one fold is used for validation, and the remaining fold is evaluated once as held-out test data. Additional cohort and cross-validation details are provided in Supplementary Section~\ref{sec:supp_gw} and Supplementary Table ~\ref{tab:s7}.

\paragraph{Quasiparticle learning and evaluation}

For each matched state,

\begin{equation}
    \Delta\varepsilon^{GW}_{n\mathbf{k}}
    =
    \varepsilon^{G_0W_0}_{n\mathbf{k}}
    -
    \varepsilon^{\mathrm{PBE}}_{n\mathbf{k}},
\end{equation}

and the quasiparticle energy is reconstructed by adding the predicted correction to the corresponding PBE baseline. State matching follows the fixed valence- and conduction-band ordering used in the benchmark.

Performance is reported as the mean absolute error over the twelve available quasiparticle states and as the error in the resulting fundamental band gap. Results are averaged across the five folds, with uncertainties reported as the sample standard deviation. To assess sensitivity to the baseline electronic structure, the same \gzwz{} target is learned separately from GPAW and FHI-aims PBE inputs \citep{Mortensen2024GPAW,Blum2009FHIaims}. State-resolved differences between the two PBE baselines are additionally evaluated after independent energy alignment to characterize where baseline inconsistency is largest across the valence and conduction manifolds. State-resolved errors across the valence and conduction manifolds are shown in Supplementary Fig.~\ref{fig:s4}, with fold-resolved state and band-gap errors reported in Supplementary Table~\ref{tab:s7}.

\paragraph{Data Availability}

All datasets used in this study will be made available upon publication.

\paragraph{Code Availability}

All code used in this study will be made available upon publication.

\paragraph{Acknowledgments}
A.A. acknowledges support in part from the Bren Endowed Chair, the Office of Naval Research through MURI grant N00014-18-1-2624, DARPA ExpMath (HR0011), and the AI2050 Senior Fellow program at Schmidt Sciences. W.A.G. acknowledges support from the Liquid Sunlight Alliance, which is supported by the U.S. Department of Energy, Office of Science, Office of Basic Energy Sciences, Fuels from Sunlight Hub under Award Number DE-SC0021266, and from the Hong Kong Quantum AI Lab, AIR@InnoHK of the Hong Kong. M.K. acknowledges support through the Caltech Summer Undergraduate Research Fellowship. B.S.K. acknowledges support from the Pritzker AI+Science Fund, DARPA Biological Technologies (HR0011), and the Eddleman Graduate Fellowship. D.K. acknowledges support from the Pritzker AI+Science Fund and DARPA Biological Technologies (HR0011). Authors acknowledge Maurice Hanisch, Abdullah Maruf, and Sergei Kliavinek for helpful discussions. Authors also acknowledge Akhil Sugathan Nair for providing the HSE06 dataset used in the work, and Jens J{\o}rgen Mortensen and Kristian Sommer Thygesen for providing the C2DB \gzwz{} dataset. This work used the Delta system at the National Center for Supercomputing Applications through allocation DMR160114 from the Advanced Cyberinfrastructure Coordination Ecosystem: Services \& Support (ACCESS) program, which is supported by National Science Foundation grants \#2138259, \#2138286, \#2138307, \#2137603, and \#2138296.

We acknowledge the use of large language models (LLMs) for initial brainstorming, assistance with handling public datasets, development of analysis pipelines around publicly available packages, and support with figure design, visualization, and figure-generation workflows. LLMs were not used in the development of the core model. All implementations, analyses, figures, and reported results were independently checked and validated by the authors.


\bibliographystyle{unsrt}
\bibliography{references}


\clearpage

\setcounter{section}{0}
\setcounter{subsection}{0}
\setcounter{figure}{0}
\setcounter{table}{0}
\setcounter{equation}{0}

\renewcommand{\thesection}{S\arabic{section}}
\renewcommand{\thesubsection}{\thesection.\arabic{subsection}}
\renewcommand{\thefigure}{S\arabic{figure}}
\renewcommand{\thetable}{S\arabic{table}}
\renewcommand{\theequation}{S\arabic{equation}}

\renewcommand{\theHsection}{S\arabic{section}}
\renewcommand{\theHsubsection}{\theHsection.\arabic{subsection}}
\renewcommand{\theHfigure}{S\arabic{figure}}
\renewcommand{\theHtable}{S\arabic{table}}
\renewcommand{\theHequation}{S\arabic{equation}}

\newgeometry{margin=21mm,includeheadfoot}

\setlist{nosep,leftmargin=1.4em}
\setlength{\parindent}{0pt}
\setlength{\parskip}{0.48em}
\setlength{\tabcolsep}{2pt}
\setlength{\LTpre}{0.4em}
\setlength{\LTpost}{0.8em}
\setlength{\emergencystretch}{3em}

\captionsetup[table]{
    font=normalsize,
    skip=8pt
}
\captionsetup[longtable]{
    font=normalsize,
    skip=8pt
}
\captionsetup[figure]{
    font=normalsize,
    skip=6pt
}

\begin{center}
{\large\bfseries Supplementary Information}\\[0.3em]
{\large Beyond DFT electronic band structures via dual-space orbital learning}
\end{center}

\vspace{1em}

\section{Dataset construction and splits}
\label{sec:supp_dataset}

The three-dimensional benchmark is derived from the paired PBE--HSE06 all-electron dataset \citep{nair2025hse06_dataset,Perdew1996PBE,Heyd2003HSE,Krukau2006HSE06}. We distinguish \emph{parent materials} from \emph{band-structure entries}: an unpolarized parent contributes one entry, whereas a collinear spin-polarized parent contributes separate spin-up and spin-down entries. All entries associated with the same parent are retained within the same data partition.

The paired source dataset contains 6,576 parent materials and 8,040 band-structure entries. To reduce cases in which the baseline and target calculations represent different magnetic states, a complete parent is removed when the absolute difference between its PBE and HSE06 magnetic moments exceeds $0.5\,\mu_{\mathrm B}$. This leaves 6,045 parents and 6,978 entries, including 933 spin-polarized parents represented by 1,866 explicit spin-channel entries. Materials reserved for the correlated-material and experimental downstream benchmarks are then excluded, yielding a final benchmark dataset of 5,988 parents and 6,900 entries.

The final benchmark dataset is partitioned at the parent-material level into approximately 8:1:1 training, validation, and test sets. Corresponding spin channels and repeated entries remain grouped within the same partition. Three independent grouped splits are used for the model/split replicates reported in the main benchmark. Exact parent and entry counts are given in Supplementary Table~\ref{tab:s1}.

\begin{table*}[!htbp]
\centering
\caption{
\textbf{Dataset composition and grouped-split balance.}
Parent-material and band-structure-entry counts are exact.
Atoms and $\mathbf{k}$-point counts are reported as entry-weighted means with
interquartile ranges shown beneath in brackets.
$k_{\mathrm{HS}}$ and $k_{\mathrm{full}}$ denote the stored high-symmetry
and full-path $\mathbf{k}$-point counts, respectively.
Electronic-character fractions are parent weighted.
All retained datasets contain 59 chemical elements.
}
\label{tab:s1}

\small
\setlength{\tabcolsep}{4pt}
\renewcommand{\arraystretch}{1.15}

\begin{tabular}{
p{0.105\textwidth}
p{0.085\textwidth}
r
r
p{0.072\textwidth}
p{0.072\textwidth}
p{0.082\textwidth}
r
r
r
}
\toprule
Stage / seed
& Partition
& Parents
& Entries
& \makecell{Atoms\\[-1pt]{\scriptsize mean}\\[-2pt]{\scriptsize [IQR]}}
& \makecell{$k_{\mathrm{HS}}$\\[-1pt]{\scriptsize mean}\\[-2pt]{\scriptsize [IQR]}}
& \makecell{$k_{\mathrm{full}}$\\[-1pt]{\scriptsize mean}\\[-2pt]{\scriptsize [IQR]}}
& $I\!\to\!I$
& $M\!\to\!I$
& $M\!\to\!M$
\\
&&&&&&&
\multicolumn{3}{c}{\scriptsize Parent fraction (\%)}
\\
\cmidrule(lr){8-10}

\midrule

\multicolumn{10}{l}{\textit{Dataset construction}} \\[2pt]

Paired source
& All
& 6576
& 8040
& \makecell{26.6\\[-2pt]{\scriptsize [9--32]}}
& \makecell{13.1\\[-2pt]{\scriptsize [12--15]}}
& \makecell{184.0\\[-2pt]{\scriptsize [131--220]}}
& 50.8
& 7.2
& 42.0
\\

After magnetic-state cleaning
& All
& 6045
& 6978
& \makecell{27.6\\[-2pt]{\scriptsize [9--34]}}
& \makecell{13.1\\[-2pt]{\scriptsize [12--15]}}
& \makecell{181.1\\[-2pt]{\scriptsize [130--216]}}
& 54.6
& 6.3
& 39.1
\\

Final benchmark dataset
& All
& 5988
& 6900
& \makecell{27.8\\[-2pt]{\scriptsize [10--34]}}
& \makecell{13.1\\[-2pt]{\scriptsize [12--15]}}
& \makecell{180.6\\[-2pt]{\scriptsize [129--216]}}
& 54.5
& 6.2
& 39.3
\\

\addlinespace[5pt]
\multicolumn{10}{l}{\textit{Grouped train/validation/test splits}} \\[2pt]

\multirow{3}{*}{Seed 42}
& Train
& 4786
& 5513
& \makecell{27.8\\[-2pt]{\scriptsize [10--34]}}
& \makecell{13.1\\[-2pt]{\scriptsize [12--15]}}
& \makecell{180.7\\[-2pt]{\scriptsize [129--216]}}
& 54.5
& 6.2
& 39.3
\\

& Val
& 600
& 694
& \makecell{27.7\\[-2pt]{\scriptsize [10--33]}}
& \makecell{13.0\\[-2pt]{\scriptsize [12--15]}}
& \makecell{177.0\\[-2pt]{\scriptsize [126--212]}}
& 54.7
& 6.3
& 39.0
\\

& Test
& 602
& 693
& \makecell{27.9\\[-2pt]{\scriptsize [10--32]}}
& \makecell{13.2\\[-2pt]{\scriptsize [12--15]}}
& \makecell{182.5\\[-2pt]{\scriptsize [130--217]}}
& 54.7
& 6.0
& 39.4
\\

\addlinespace[3pt]

\multirow{3}{*}{Seed 43}
& Train
& 4786
& 5520
& \makecell{27.8\\[-2pt]{\scriptsize [10--34]}}
& \makecell{13.1\\[-2pt]{\scriptsize [12--15]}}
& \makecell{180.1\\[-2pt]{\scriptsize [129--215]}}
& 54.5
& 6.2
& 39.4
\\

& Val
& 598
& 684
& \makecell{27.4\\[-2pt]{\scriptsize [9--34]}}
& \makecell{13.1\\[-2pt]{\scriptsize [12--15]}}
& \makecell{182.8\\[-2pt]{\scriptsize [131--218]}}
& 54.7
& 6.7
& 38.6
\\

& Test
& 604
& 696
& \makecell{27.8\\[-2pt]{\scriptsize [10--34]}}
& \makecell{13.2\\[-2pt]{\scriptsize [12--15]}}
& \makecell{182.0\\[-2pt]{\scriptsize [129--216]}}
& 54.6
& 6.1
& 39.2
\\

\addlinespace[3pt]

\multirow{3}{*}{Seed 44}
& Train
& 4787
& 5518
& \makecell{27.7\\[-2pt]{\scriptsize [10--34]}}
& \makecell{13.1\\[-2pt]{\scriptsize [12--15]}}
& \makecell{180.5\\[-2pt]{\scriptsize [129--216]}}
& 54.5
& 6.2
& 39.3
\\

& Val
& 606
& 706
& \makecell{27.1\\[-2pt]{\scriptsize [9--32]}}
& \makecell{13.0\\[-2pt]{\scriptsize [11--14]}}
& \makecell{183.2\\[-2pt]{\scriptsize [130--216]}}
& 54.3
& 6.4
& 39.3
\\

& Test
& 595
& 676
& \makecell{29.2\\[-2pt]{\scriptsize [10--32]}}
& \makecell{13.2\\[-2pt]{\scriptsize [12--15]}}
& \makecell{178.2\\[-2pt]{\scriptsize [129--212]}}
& 54.8
& 5.9
& 39.3
\\

\bottomrule
\end{tabular}
\end{table*}

\clearpage
\section{Orbital representation, architecture, and training}
\label{sec:supp_model}

\paragraph{Orbital representation}

For each PBE calculation, \ourmodel{} uses atom-centered blocks of the converged Kohn--Sham Hamiltonian $\mathbf H$, overlap matrix $\mathbf S$, density matrix $\mathbf P$, and spin-density matrix
$\mathbf P_{\mathrm{spin}}^{(\sigma)}
=
\mathbf P_{\sigma}-\mathbf P_{\bar{\sigma}}.$

For non-spin-polarized entries, the spin-density channel is zero. The input uses a trimmed-valence atomic-orbital representation with one-center and pair blocks padded to a maximum dimension of 16. Real-space orbital blocks describe local electronic interactions, while the corresponding reciprocal-space matrices provide explicit $\mathbf{k}$-resolved information. The production model uses the real components of $\mathbf H(\mathbf{k})$, $\mathbf S(\mathbf{k})$, $\mathbf P(\mathbf{k})$, and $\mathbf P_{\mathrm{spin}}(\mathbf{k})$ with imaginary reciprocal channels disabled.

Crystal-momentum dependence is additionally introduced through Bloch-phase features on each periodic pair,
$
\left[
\sin(\mathbf{k}\cdot\mathbf r_{ij\mathbf R}),
\cos(\mathbf{k}\cdot\mathbf r_{ij\mathbf R})
\right]$, 
$
\mathbf r_{ij\mathbf R}=\mathbf r_j-\mathbf r_i+\mathbf R .
$
The reciprocal orbital matrices are obtained directly from the electronic-structure exports and are not reconstructed from the truncated graph.

\paragraph{Model architecture}

\ourmodel{} consists of two jointly trained periodic equivariant branches. The affine branch uses real-space orbital information and Bloch-phase conditioning to predict a band-dependent correction to the PBE dispersion, while the dual-space residual branch additionally uses explicit reciprocal-space orbital matrices to predict a $\mathbf{k}$-dependent residual. In the final configuration, the affine branch inherits four interaction and four decoding layers from the model defaults, whereas the residual branch explicitly uses two interaction and two decoding layers. The residual is projected on the high-symmetry reference set to remove its constant and PBE-linear components before being added to the affine correction.

Electronic states are addressed through signed band queries relative to the band edges, with negative indices for valence bands and positive indices for conduction bands. The affine decoder additionally receives five deterministic PBE band descriptors: the band-edge-referenced band mean, standard deviation, minimum, maximum, and band width, as defined in the main Methods. These descriptors are computed over the valid high-symmetry-point subset and use fixed energy scales of 10~eV for the mean, minimum, and maximum terms and 5~eV for the standard deviation and width. The signed-band representation allows the model to accommodate variable numbers of valence and conduction states rather than requiring a fixed-dimensional band output.

\paragraph{Training}

The primary target is the state-resolved PBE-to-HSE06 correction. Training uses only the stored high-symmetry $\mathbf{k}$-points. The primary state objective is an energy-weighted SmoothL1 loss with transition parameter $\beta=0.1$~eV. An additional band-offset term penalizes the SmoothL1 error of the mean signed residual for each material and band over the supervised high-symmetry points, and an indirect-gap term penalizes errors in the reconstructed fundamental gap. The final objective is
\[
\mathcal{L}
=
\mathcal{L}_{\mathrm{state}}
+
\mathcal{L}_{\mathrm{offset}}
+
0.2\,\mathcal{L}_{\mathrm{gap}}.
\]

All other auxiliary objectives, including direct-gap, dispersion, curvature, ranking, and effective-mass terms, have zero weight in the production model.

State weights decay with the PBE energy relative to the band-edge reference using a characteristic scale of $10$~eV. Models are trained for 100 epochs with AdamW, a 20-epoch linear warm-up and cosine learning-rate decay. The reported checkpoints are selected by the lowest unweighted mean validation MAE over the 2, 5, 10, and 20~eV energy windows. Exact architectural and optimization settings are summarized in Supplementary Table~\ref{tab:s2}.

\begin{table}[!htbp]
\centering
\caption{Architecture and training settings for the three-dimensional PBE-to-HSE06 \ourmodel{} model.}
\label{tab:s2}
\small
\begin{tabular}{lll}
\toprule
Category & Setting & Value \\
\midrule

Model
& Trainable parameters
& 8,940,709 \\

& Branches
& Real+phase affine; dual-space residual \\

& Interaction / decoder layers
& affine 4 / 4; residual 2 / 2 \\

\addlinespace
Representation
& Orbital operators
& $H$, $S$, $P$, $P_{\mathrm{spin}}$ \\

& Reciprocal representation
& Real components of orbital matrices \\

& AO block dimension
& 16 \\

& Radial basis functions
& 16 \\

& Attention
& 4 heads; dimension 64 per head \\

\addlinespace
Decoder
& Material/$\mathbf{k}$ latent dimension
& 64 \\

& Signed-band embedding dimension
& 32 \\

& Band descriptors
& mean, s.d., minimum, maximum, width \\

& Residual projection
& HS constant and PBE-affine components removed \\

\addlinespace
Training
& Target
& $E^{\mathrm{HSE06}}-E^{\mathrm{PBE}}$ \\

& Supervised $\mathbf{k}$-points
& High-symmetry points only \\

& Loss
& SmoothL1 state + offset + $0.2\times$ indirect-gap loss \\

& Optimizer
& AdamW, $5\times10^{-4}$ initial learning rate \\

& Weight decay / betas
& $0.005$ / $(0.9,0.95)$ \\

& Schedule
& 20-epoch warm-up; cosine decay; 100 epochs \\

& Precision
& bfloat16 autocast \\

& Checkpoint selection
& Mean validation MAE over 2, 5, 10 and 20 eV windows \\

\bottomrule
\end{tabular}
\end{table}

\begin{algorithm}[t]
\caption{O-kRystal training and evaluation procedure}
\KwIn{Parent-grouped $\mathcal{D}_{\mathrm{train}}$,
$\mathcal{D}_{\mathrm{val}}$, and $\mathcal{D}_{\mathrm{test}}$}
\KwOut{Selected model $f_{\theta^*}$ and test predictions}

\textit{Notation:}
$\mathcal{O}^{R}$ and $\mathcal{O}^{k}$ are real- and reciprocal-space
orbital operators; $\boldsymbol{\phi}$ denotes Bloch phases; $q$ is the
signed-band query; and $\mathbf{d}$ denotes PBE descriptors.\;

\textbf{1. Construct orbital features and targets}\;
\For{$i \in \mathcal{D}_{\mathrm{train}}$}{
  $\mathbf{T}_i \gets
  (\mathcal{O}^{R}_i,\mathcal{O}^{k}_i,\boldsymbol{\phi}_i,q_i,\mathbf{d}_i)$\;
  $\Delta_i \gets
  \mathbf{E}^{\mathrm{HSE06}}_i-\mathbf{E}^{\mathrm{PBE}}_i$\;
}
\For{$i \in \mathcal{D}_{\mathrm{val}}\cup\mathcal{D}_{\mathrm{test}}$}{
  $\mathbf{T}_i \gets
  (\mathcal{O}^{R}_i,\mathcal{O}^{k}_i,\boldsymbol{\phi}_i,q_i,\mathbf{d}_i)$\;
}

\textbf{2. Train and select model}\;
$\{\theta_e\} \gets
\operatorname{Train}(f_\theta,\mathcal{D}_{\mathrm{train}})$\;
$\mathcal{W} \gets \{2,5,10,20\}\,\mathrm{eV}$\;
$\theta^* \gets \displaystyle\arg\min_{\theta_e}
\frac{1}{|\mathcal{W}|}\sum_{W\in\mathcal{W}}
\operatorname{MAE}_{W}(\mathcal{D}_{\mathrm{val}};\theta_e)$\;

\textbf{3. Predict test bands}\;
$\widehat{\Delta}_i \gets f_{\theta^*}(\mathbf{T}_i)$,
$i\in\mathcal{D}_{\mathrm{test}}$\;
$\widehat{\mathbf{E}}^{\mathrm{HSE06}}_i \gets
\mathbf{E}^{\mathrm{PBE}}_i+\widehat{\Delta}_i$\;

\end{algorithm}

\paragraph{Comparison models and representation ablation}

The scissor and Linear+KRR baselines operate directly on the PBE band energies and derived state descriptors and contain no learned geometric or orbital representation. The scissor model fits separate global mean PBE-to-HSE06 corrections for valence- and conduction-side states using the training set only. Linear+KRR uses six standardized PBE-derived descriptors in a ridge-linear model followed by RBF kernel-ridge regression of the remaining residual.

The Geometry GNN provides a neural control in which the orbital matrices are removed while retaining the PBE band baseline and explicit crystal-momentum conditioning. It uses atomic-number embeddings, periodic geometric pair features, and Bloch-phase information in a six-layer gated message-passing network with a PBE-conditioned affine band decoder. The Real+Bloch-phase ablation instead retains the real-space orbital representation of \ourmodel{} together with Bloch-phase conditioning, while omitting the explicit reciprocal-space orbital matrices.

\begin{table}[!htbp]
\centering
\caption{
\textbf{Definitions and implementation details for comparison models and
representation controls.}
For the neural learning-curve comparison, Geometry GNN, Real+Bloch phase, and \ourmodel{} use matched seed-specific training subsets, the same final training objective and optimization schedule, raw \texttt{best\_avg\_win} checkpoints, and the same high-symmetry-point evaluation protocol. The three neural models therefore differ primarily in their electronic representation and architecture.
}
\label{tab:s3}
\small
\begin{tabular}{p{0.17\textwidth}p{0.20\textwidth}p{0.54\textwidth}}
\toprule
Model & Component & Setting \\
\midrule

Scissor
& Inputs
& PBE band energies and stored occupied/unoccupied assignment \\

& Fit
& Separate global mean PBE-to-HSE06 corrections for occupied and unoccupied training states; no validation or test fitting \\

\addlinespace

Linear+KRR
& Inputs
& Relative PBE energy, occupation indicator, band width, band standard deviation, $\log(1+\text{nearest-band spacing})$, and $\log(N_{\mathrm{atoms}})$ \\

& Linear stage
& Training-set standardization followed by ridge regression with $\lambda_{\mathrm{ridge}}=10^{-3}N$, where $N$ is the number of sampled training states \\

& Kernel stage
& RBF kernel-ridge regression of the linear-model residual with $\gamma=1/6$ and regularized system $K^{\mathsf T}K + 10^{-2}N K_{cc} + 10^{-6}I$ \\

& Sampling
& Up to 768 high-symmetry states per training entry; up to 768 kernel centres sampled without replacement from the pooled training states \\

\addlinespace

Geometry GNN
& Inputs
& Atomic species, periodic geometry, Bloch-phase features, PBE band energies, signed-band queries, and five PBE band descriptors; no orbital matrices \\

& Architecture
& Six gated periodic message-passing layers with hidden dimension 256;
adaptive nearest-32 periodic graph initialized at 12 bohr and expanded to at
most 32 bohr; 16 Gaussian radial functions over 0--32 bohr \\

& Decoder
& PBE-conditioned affine band decoder with 32-dimensional signed-band
embedding and 64-dimensional band latent \\

& Parameters
& 5,291,356 \\

\addlinespace

Real+Bloch phase
& Inputs
& Real-space $H$, $S$, $P$, and $P_{\mathrm{spin}}$ orbital blocks,
periodic geometry, Bloch-phase features, PBE band energies and descriptors,
and signed-band queries \\

& Architecture
& Two interaction and two decoding layers with the PBE-conditioned affine
band decoder; explicit reciprocal-space orbital matrices and reciprocal
residual branch omitted \\

& Parameters
& 3,238,604 \\

\addlinespace
\midrule

\multicolumn{2}{l}{Training}
& SmoothL1 ($\beta=0.1$) state loss with normalized energy weighting
($E_w=10$~eV), high-symmetry offset loss (weight 1), and indirect-gap loss
(weight 0.2) \\

\multicolumn{2}{l}{Optimization}
& AdamW, learning rate $5\times10^{-4}$, weight decay 0.005,
$(\beta_1,\beta_2)=(0.9,0.95)$; 20-epoch linear warm-up followed by cosine
decay; 100 epochs \\

\multicolumn{2}{l}{Checkpoint / evaluation}
& Lowest unweighted mean validation MAE over the 2, 5, 10, and 20~eV windows;
raw model weights evaluated on the corresponding retained high-symmetry
test states \\

\bottomrule
\end{tabular}
\end{table}

\clearpage
\section{Contribution of reciprocal-space orbital information}
\label{sec:supp_reciprocal_representation}

\begin{figure}[!htbp]
\centering
\includegraphics[width=\textwidth]{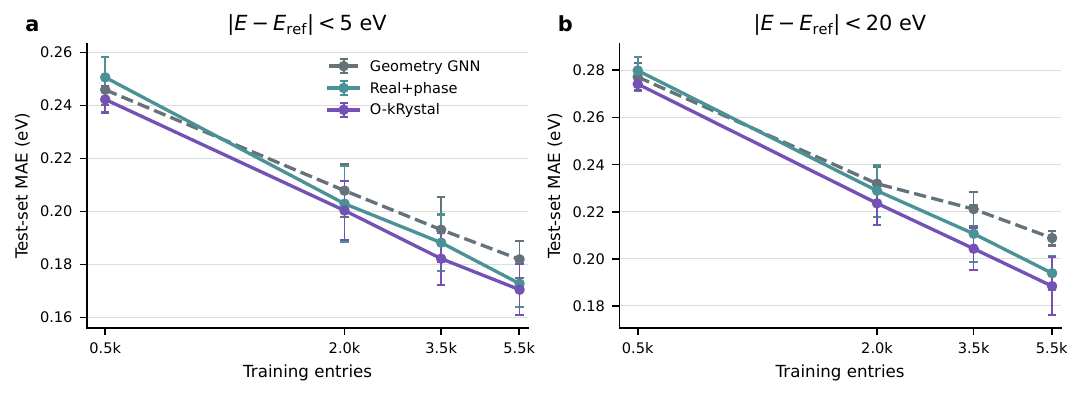}
\caption{
\textbf{Learning-curve comparison of geometric, real-space orbital, and dual-space orbital representations.}
Test-set band-energy MAE as a function of the number of training entries for states within \textbf{a}, 5~eV and \textbf{b}, 20~eV of the PBE reference energy. The Geometry GNN uses periodic crystal geometry, Bloch-phase conditioning, and the PBE band baseline and descriptors, but no orbital matrices. Real+phase additionally uses real-space orbital blocks with Bloch-phase conditioning, while \ourmodel{} further incorporates explicit reciprocal-space orbital matrices. All models are evaluated on the corresponding fixed test sets using matched training-data subsets. Points and error bars denote the mean and sample standard deviation across seeds 42--44. The representations perform similarly in the low-data regime, whereas the full dual-space model shows a progressively clearer advantage with increasing training-set size, particularly across the broader 20~eV energy window.
}
\label{fig:s1}
\end{figure}

To examine the contribution of orbital representation, we compare the full dual-space \ourmodel{} with two controls across matched training-set sizes (Supplementary Fig.~\ref{fig:s1}). The Geometry GNN retains the PBE band baseline, band descriptors, periodic geometry, and Bloch-phase conditioning but removes the orbital matrices, whereas the Real+phase model retains the real-space orbital blocks and Bloch-phase features while omitting the explicit reciprocal-space orbital matrices. The Geometry GNN uses a distance-selected periodic graph, whereas Real+phase and O-kRystal share the same electronically preselected real-space orbital graph. Their comparison therefore evaluates the complete dual-space architecture against a real-space orbital alternative that omits the explicit reciprocal-space matrices.

The three representations perform similarly in the smallest-data regime, but their learning curves separate as the training set increases. Real-space orbital information improves upon the geometry-based representation, while the full dual-space model achieves the lowest error at the largest training size in both energy windows. The separation is more pronounced over the 20~eV window, where accurate correction requires representing a broader range of electronic states. These results show that the complete dual-space model achieves the lowest error in the larger-data regime, with the separation becoming clearer over the broader energy window.

\clearpage
\section{Reciprocal-space generalization}
\label{sec:supp_k_generalization}

\ourmodel{} is trained only on stored high-symmetry $\mathbf{k}$-points for the three-dimensional PBE-to-HSE06 task. Generalization is evaluated at unseen points along the corresponding high-symmetry paths using the same state-matching and energy-window conventions as the primary benchmark. Because predictions are conditioned on the supplied $\mathbf{k}$-point set, path-based evaluations use the complete query set for each path.

\begin{figure}[!htbp]
\centering
\includegraphics[width=\textwidth]{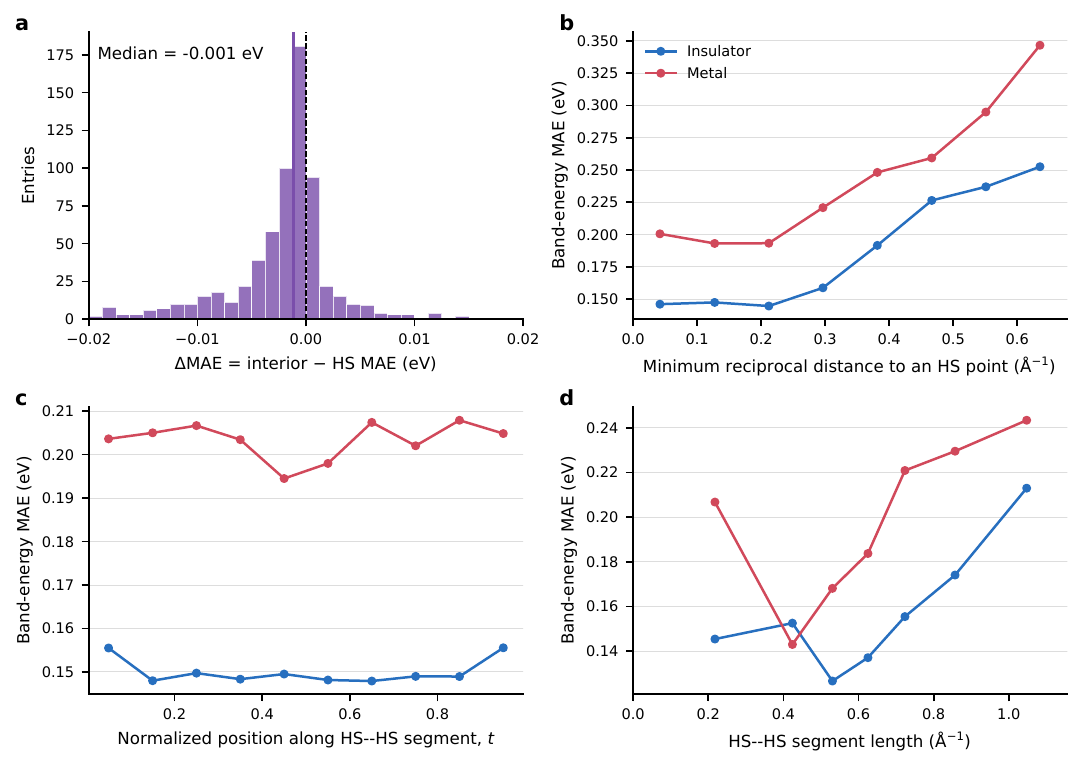}
\caption{
\textbf{Dependence of unseen-point error on reciprocal-space position.} Diagnostics for path-interior states within $|E-\eref|<5$~eV using the seed-42 raw-weight checkpoint.
\textbf{a}, Distribution of the per-entry difference between interior-point and high-symmetry-point MAE.
\textbf{b}, Band-energy MAE as a function of reciprocal distance to the nearest high-symmetry point.
\textbf{c}, Band-energy MAE as a function of normalized position within a path segment.
\textbf{d}, Band-energy MAE as a function of segment length. Panels b--d pool valid states across k-point rows and are therefore state-weighted rather than material-balanced. The elevated metallic bins are concentrated in a small number of high-error materials and should not be interpreted as a universal monotonic dependence on reciprocal-space distance or segment length.
}
\label{fig:s2}
\end{figure}

Supplementary Fig.~\ref{fig:s2} examines whether the error at unseen path-interior points depends systematically on reciprocal-space location. The distribution of the difference between interior-point and high-symmetry-point MAE is centered near zero, consistent with the nearly unchanged aggregate error reported in the main text. For insulating systems, the interior-point error increases gradually with reciprocal distance from the nearest supervised high-symmetry point, while remaining nearly independent of normalized position along the path segment. A similar increase is observed for longer high-symmetry-to-high-symmetry segments. Metallic systems exhibit larger fluctuations, with the highest-error bins dominated by a small number of difficult materials, primarily Ti$_2$P and Zr$_2$P, rather than by a uniform trend across the metallic subset.

Representative full-path predictions are shown in
Supplementary Fig.~\ref{fig:s3}.

\begin{figure}[!htbp]
\centering
\includegraphics[width=\textwidth]{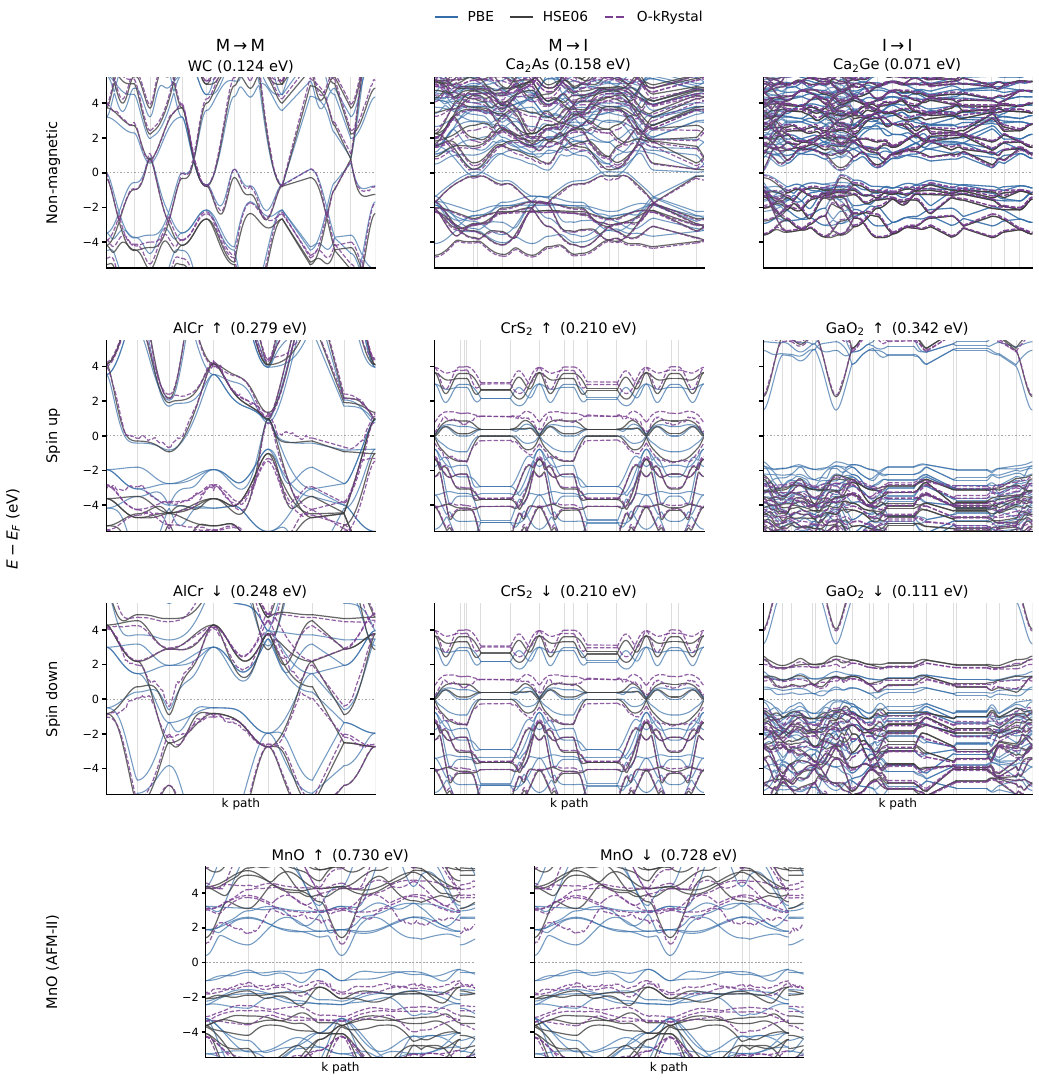}
\caption{
\textbf{Representative full-path band-structure predictions.}
PBE, HSE06, and \ourmodel{} band energies are shown along identical stored $\mathbf{k}$-paths for representative materials spanning insulating, metallic, and spin-polarized electronic structures.
The gallery includes GaO$_2$ and WC, whose dense Brillouin-zone DOS reconstructions are shown in Figs.~\ref{fig:overview}b and \ref{fig:generalization}c, respectively, together with AFM-II MnO, whose orbital-resolved spectrum is shown in Fig.~\ref{fig:overview}c.
Additional examples illustrate performance across different PBE-to-HSE06 electronic-character changes and spin configurations.
Values in parentheses denote the \ourmodel{} band-energy MAE for each displayed entry within $\lvert E-\eref\rvert < 5$~eV.
}
\label{fig:s3}
\end{figure}

\clearpage
\section{Strongly correlated materials}
\label{sec:supp_correlated}

\paragraph{Benchmark construction}

The correlated-material benchmark is constructed independently of model performance.
Candidate materials are retained only when the represented magnetic and electronic state is consistent with the corresponding literature-supported phase and when primary literature supports classification as a strongly correlated, Mott-like, charge-transfer, or correlated-metal system.
Applying these criteria yields six materials: CrO$_2$, LaCoO$_3$, LaNiO$_3$, AFM-II MnO, AFM-II NiO, and YTiO$_3$ \citep{Tsujioka1997CrO2,Korotin1998CrO2,Saitoh1997LaCoO3,Haverkort2006LaCoO3,Ouellette2010LaNiO3,Karner2019LaNiO3,Roth1958Monoxides,Liu2020Monoxides,Klotz2020MnO,Kunes2007NiO,Morikawa1996YTiO3,Iga2004YTiO3}.
All six materials are excluded before construction of the primary training, validation, and test splits.
Candidate systems are omitted when the available calculation does not provide a defensible representation of the literature-supported phase, including cases with incompatible magnetic ordering, insufficient cell size for the reported antiferromagnetic state, or an HSE06 solution inconsistent with the established electronic character.

The quantitative analysis uses the representative seed-42 model evaluated on dense, material-specific Brillouin-zone meshes.
Band-energy errors are computed over states selected by inclusive PBE-centered energy windows.
No reciprocal-space quadrature weights are applied to the band-energy MAE.
For spin-polarized systems, retained states from both spin channels are pooled within each parent material before calculating its MAE, after which the six parent materials are weighted equally.
The primary comparison uses the 5~eV window, while the 20~eV window provides a broader-energy-range evaluation.

\begin{table}[ht!]
\centering
\caption{\textbf{Correlated-material band-energy benchmark.}
Band-energy mean absolute errors (MAEs) are evaluated using the representative seed-42 model on dense Brillouin-zone meshes.
States are selected using inclusive PBE-centered energy windows of 5 and 20~eV.
For spin-polarized systems, retained states are pooled across spin channels before calculating the parent-material MAE.
The final row reports the equal-material mean across the six materials.
All values are in eV.}
\label{tab:corr_band_mae}
\small
\begin{tabular}{lcccccc}
\toprule
& \multicolumn{3}{c}{$\leq 5$~eV} 
& \multicolumn{3}{c}{$\leq 20$~eV} \\
\cmidrule(lr){2-4}
\cmidrule(lr){5-7}
Material
& PBE
& Geometry GNN
& \ourmodel{}
& PBE
& Geometry GNN
& \ourmodel{} \\
\midrule
CrO$_2$     & 0.974 & 0.348 & \textbf{0.279} & 1.259 & 0.440 & \textbf{0.332} \\
LaCoO$_3$   & 1.741 & 0.997 & \textbf{0.591} & 1.600 & 0.956 & \textbf{0.516} \\
LaNiO$_3$   & 1.576 & 0.462 & \textbf{0.367} & 1.469 & \textbf{0.364} & 0.368 \\
YTiO$_3$    & 1.036 & 0.198 & \textbf{0.102} & 1.201 & 0.177 & \textbf{0.121} \\
MnO         & 1.678 & 0.692 & \textbf{0.605} & 1.537 & 0.500 & \textbf{0.489} \\
NiO         & 2.186 & 1.455 & \textbf{0.930} & 1.766 & 1.046 & \textbf{0.632} \\
\midrule
Material-balanced mean
            & 1.532 & 0.692 & \textbf{0.479}
            & 1.472 & 0.581 & \textbf{0.410} \\
\bottomrule
\end{tabular}
\end{table}

Within the 5~eV window, \ourmodel{} reduces the equal-material mean band-energy MAE from 1.532~eV for PBE and 0.692~eV for Geometry GNN to 0.479~eV.
The same aggregate ordering is retained over the wider 20~eV window, where the corresponding MAEs are 1.472, 0.581, and 0.410~eV, respectively.
At the material level, \ourmodel{} gives the lowest 5~eV error for all six systems.
At 20~eV, Geometry GNN and \ourmodel{} perform similarly for LaNiO$_3$, while \ourmodel{} retains the lower aggregate error.

\paragraph{Fundamental-gap recovery}

Fundamental gaps are evaluated over the dense Brillouin-zone predictions using the stored valence--conduction band boundary.
For each spin channel, the gap is defined as the minimum conduction-band energy minus the maximum valence-band energy.
For spin-polarized materials, the material-level gap is the minimum over spin channels.
Negative gaps are classified as metallic and positive gaps as insulating.

\begin{table}[ht!]
\centering
\caption{\textbf{Fundamental-gap recovery for the correlated-material benchmark.}
Predicted fundamental gaps are compared with the corresponding HSE06 values.
Values in parentheses denote the absolute gap error relative to HSE06.
Negative gaps indicate metallic electronic character and positive gaps indicate insulating character.
All values are in eV.}
\label{tab:corr_gap}
\small
\begin{tabular}{lcccc}
\toprule
Material
& HSE06
& PBE
& Geometry GNN
& \ourmodel{} \\
\midrule
CrO$_2$
& $-0.809$
& $-1.162$ (0.353)
& $+1.052$ (1.861)
& $+1.692$ (2.501) \\

LaCoO$_3$
& $+2.669$
& $-0.084$ (2.753)
& $+0.735$ (1.934)
& $+1.852$ (0.818) \\

LaNiO$_3$
& $-2.159$
& $-1.311$ (0.849)
& $-0.744$ (1.416)
& $-0.175$ (1.985) \\

YTiO$_3$
& $+1.852$
& $-0.364$ (2.216)
& $+0.340$ (1.513)
& $+0.536$ (1.316) \\

MnO
& $+2.864$
& $+0.803$ (2.061)
& $+1.835$ (1.029)
& $+2.611$ (0.253) \\

NiO
& $+4.458$
& $+0.973$ (3.485)
& $+1.595$ (2.863)
& $+2.109$ (2.349) \\
\midrule
Gap MAE
& --
& 1.953
& 1.769
& \textbf{1.537} \\

Correct electronic character
& --
& 4/6
& 5/6
& 5/6 \\
\bottomrule
\end{tabular}
\end{table}

Across the six materials, the fundamental-gap MAE decreases from 1.953~eV for PBE to 1.769~eV for Geometry GNN and 1.537~eV for \ourmodel{}.
The HSE06 metallic or insulating character is recovered for four of six materials with PBE and five of six with both neural correction models.
For \ourmodel{}, CrO$_2$ is the remaining classification failure, with the metallic HSE06 reference predicted as insulating.

\paragraph{Orbital-resolved projected density of states}

Orbital-resolved spectral reconstruction is evaluated for transition-metal $3d$ and oxygen $2p$ channels across all six correlated materials.
To isolate differences in state-energy placement, the PBE, Geometry GNN, \ourmodel{}, and HSE06 projected densities of states are constructed using the same Mulliken orbital-projection weights obtained from the baseline PBE wavefunctions.
Only the state energies entering the reconstruction differ between methods.
Projected densities of states are evaluated on dense reciprocal-space meshes using linear tetrahedron integration followed by Gaussian broadening with $\sigma=0.10$~eV.

Agreement with HSE06 is quantified using the normalized $L_1$ error defined in Methods.
The integration window is centered at $E_{\mathrm{mid}}$ and extends from $-5$ to $+5$~eV.
Because the orbital weights are fixed across all spectra, the comparison isolates differences in orbital-resolved state-energy placement rather than changes in target-theory wavefunctions or hybridization.

\begin{table}[ht!]
\centering
\caption{\textbf{Orbital-resolved PDOS reconstruction errors for the correlated-material benchmark.}
Normalized $L_1$ errors are evaluated relative to HSE06 within $E_{\mathrm{mid}}\pm5$~eV.
All spectra use identical PBE Mulliken projection weights, such that only the state energies differ between PBE, Geometry GNN, \ourmodel{}, and HSE06.
The final row reports the equal-material mean across the six materials.
The metric is dimensionless.}
\label{tab:corr_pdos}
\small
\begin{tabular}{lcccccc}
\toprule
& \multicolumn{3}{c}{Transition-metal $3d$}
& \multicolumn{3}{c}{Oxygen $2p$} \\
\cmidrule(lr){2-4}
\cmidrule(lr){5-7}
Material
& PBE
& Geometry GNN
& \ourmodel{}
& PBE
& Geometry GNN
& \ourmodel{} \\
\midrule
CrO$_2$
& 1.055 & 0.466 & \textbf{0.456}
& 0.642 & \textbf{0.323} & 0.453 \\

LaCoO$_3$
& 1.181 & 0.839 & \textbf{0.571}
& 0.648 & 0.734 & \textbf{0.621} \\

LaNiO$_3$
& 0.931 & 0.862 & \textbf{0.634}
& 0.801 & \textbf{0.671} & 0.726 \\

YTiO$_3$
& 1.477 & 0.780 & \textbf{0.450}
& 1.309 & 0.396 & \textbf{0.255} \\

MnO
& 1.938 & 1.163 & \textbf{0.715}
& 1.378 & 1.109 & \textbf{0.676} \\

NiO
& 2.123 & 1.827 & \textbf{1.430}
& 1.440 & 1.286 & \textbf{1.056} \\
\midrule
Equal-material mean
& 1.451 & 0.989 & \textbf{0.709}
& 1.036 & 0.753 & \textbf{0.631} \\
\bottomrule
\end{tabular}
\end{table}

For transition-metal $3d$ states, \ourmodel{} lowers the normalized PDOS error relative to Geometry GNN for all six materials and reduces the equal-material mean from 0.989 to 0.709.
For oxygen $2p$ states, the aggregate error decreases from 0.753 to 0.631, although the material-resolved improvement is less uniform.
In particular, Geometry GNN gives lower oxygen $2p$ errors for CrO$_2$ and LaNiO$_3$, whereas \ourmodel{} performs better for the remaining four materials.
The stronger and more consistent improvement in the transition-metal $3d$ channel motivates the orbital-resolved analysis discussed in the main text.

\paragraph{Direct-source MnO case study}

The MnO spectrum displayed in Fig.~\ref{fig:overview}c is generated from a separately retained direct-source evaluation and is not included in the rebuilt six-material PDOS aggregate reported in Table~\ref{tab:corr_pdos}.
The calculation represents rocksalt MnO in the AFM-II magnetic configuration along $[111]$ using the corresponding four-atom rhombohedral cell \citep{Roth1958Monoxides,Klotz2020MnO}.
The PBE and HSE06 calculations use the same resolved $12\times12\times12$ reciprocal-space mesh.
The orbital-resolved reconstruction follows the same fixed-projection convention used in the six-material benchmark: PBE Mulliken weights are retained while the state energies are replaced by the corresponding \ourmodel{}-predicted HSE06 values.
The direct-source spectrum is used as an illustrative case study of the redistribution of Mn $3d$ and O $2p$ states and is kept numerically separate from the rebuilt benchmark.

\clearpage
\section{Experimental gap and effective-mass benchmark}
\label{sec:supp_mstar60}

The experimental benchmark is derived from the mstar60 dataset \cite{laurien2022mstar60_dataset} and restricted to bulk materials for which the required experimental, PBE, and HSE06 quantities are available. The final cohort contains 31 directional effective-mass observations across ten materials, grouped into III--V semiconductors (GaAs, GaN, InP), chalcogenides (CdS, CdTe, bulk 2H-MoS$_2$, bulk 2H-WS$_2$), and other semiconductors (Si, SiC, BN). Monolayers are excluded from the bulk comparison, Pb chalcogenides are treated as out of domain because Pb is absent from the training chemistry, and black phosphorus is excluded because the required PBE effective masses are unavailable. One GaAs observation with a non-positive PBE mass and one CdTe observation with an unreliable HSE06 reference value are also excluded.

Because the available HSE06 supervision does not densely resolve the local curvature around band extrema, effective masses are not predicted directly from \ourmodel{} band dispersions. Instead, we use a gap-informed reconstruction motivated by the Kane model, which relates changes in band-edge effective mass to the electronic band gap through interband coupling \citep{Kane1957Band,Lin2022Kane}. Specifically, the baseline PBE effective mass is rescaled according to the ratio between the \ourmodel{}-predicted and PBE fundamental gaps. The resulting relation is a calibrated generalization of this gap--mass dependence rather than a direct implementation of the Kane Hamiltonian.

For this analysis, the PBE and predicted HSE06 fundamental gaps are evaluated over the complete stored band path as the conduction-band minimum minus the valence-band maximum. Reconstruction parameters are calibrated separately for each held-out material family using HSE06 effective masses from the remaining families. Experimental effective masses are used only for final evaluation and do not enter the calibration.

For observation $i$ belonging to held-out family $f$, the reconstruction uses
\begin{equation}
\beta_{\mathrm{raw},i}
=
1-\delta_f m_i^{\mathrm{PBE}},
\end{equation}
where $m_i^{\mathrm{PBE}}$ is expressed in units of the free-electron mass.

The exponent is regularized as
\begin{equation}
\beta_i =
\begin{cases}
\min(\beta_{\mathrm{raw},i},1),
& \beta_{\mathrm{raw},i}\geq0,\\[3pt]
-\beta_{\mathrm{floor},f}
\tanh\!\left[
\frac{-\beta_{\mathrm{raw},i}}
{\max(\beta_{\mathrm{floor},f},10^{-3})}
\right],
& \beta_{\mathrm{raw},i}<0.
\end{cases}
\end{equation}

The gap ratio is
\begin{equation}
r_i =
\min\!\left[
10^3,
\frac{\max(E_{g,i}^{\ourmodel{}},10^{-3}\,\mathrm{eV})}
{\max(E_{g,i}^{\mathrm{PBE}},10^{-3}\,\mathrm{eV})}
\right],
\end{equation}
and the reconstructed effective mass is
\begin{equation}
m_i^{\ourmodel{}^{*}}
=
m_i^{\mathrm{PBE}} r_i^{\beta_i}.
\end{equation}

The parameters $(\delta_f,\beta_{\mathrm{floor},f})$ are selected by exhaustive grid search to minimize the material-balanced mean absolute logarithmic error relative to HSE06 effective masses in the calibration families. The search uses $\delta=0.1,0.2,\ldots,11.6$ and $\beta_{\mathrm{floor}}=0.05,0.10,\ldots,1.00$. The resulting parameters are $(4.4,0.50)$ for the III--V holdout, $(4.0,0.25)$ for the chalcogenide holdout, and $(2.9,0.50)$ for the other-semiconductor holdout.

Supplementary Table~\ref{tab:s5} summarizes the benchmark cohort, and Supplementary Table~\ref{tab:s6} reports all 31 effective-mass observations used in the primary analysis.

\clearpage
\begin{table}[!htbp]
\centering
\caption{
\textbf{Effective-mass benchmark cohort.}
The primary benchmark contains 31 directional effective-mass observations
across ten bulk materials.
}
\label{tab:s5}
\small
\begin{tabular}{llrr}
\toprule
Family & Materials & Materials & Observations \\
\midrule

III--V
& GaAs, GaN, InP
& 3
& 13 \\

Chalcogenides
& CdS, CdTe, 2H-MoS$_2$, 2H-WS$_2$
& 4
& 9 \\

Other semiconductors
& Si, SiC, BN
& 3
& 9 \\

\midrule
Total
& --
& 10
& 31 \\

\bottomrule
\end{tabular}
\end{table}

\scriptsize
\begin{longtable}{
>{\raggedright\arraybackslash}p{0.08\textwidth}
>{\raggedright\arraybackslash}p{0.11\textwidth}
>{\raggedright\arraybackslash}p{0.10\textwidth}
>{\raggedright\arraybackslash}p{0.08\textwidth}
>{\raggedright\arraybackslash}p{0.15\textwidth}
>{\raggedleft\arraybackslash}p{0.08\textwidth}
>{\raggedleft\arraybackslash}p{0.07\textwidth}
>{\raggedleft\arraybackslash}p{0.07\textwidth}
>{\raggedleft\arraybackslash}p{0.09\textwidth}
}
\caption{
\textbf{Effective-mass observations used in the primary bulk benchmark.}
All masses are reported in units of the free-electron mass. \ourmodel{}$^{*}$ denotes the family-held-out gap-informed reconstruction obtained from the PBE effective mass and the fundamental gap predicted by the raw seed-42 \ourmodel{} checkpoint. It is not a direct effective-mass output
of the model.
}
\label{tab:s6}\\

\toprule
Material & Family & Mass type & $k$ point & Direction &
Experiment & PBE & HSE06 & \ourmodel{}$^{*}$ \\
\midrule
\endfirsthead

\toprule
Material & Family & Mass type & $k$ point & Direction &
Experiment & PBE & HSE06 & \ourmodel{}$^{*}$ \\
\midrule
\endhead

GaAs & III--V & $m_n$ & $\Gamma$ & -- & 0.066 & 0.028 & 0.062 & 0.050 \\
GaAs & III--V & $m_{n,\perp}$ & $X_6$ & -- & 0.230 & 0.233 & 0.223 & 0.229 \\
GaAs & III--V & $m_{n,\perp}$ & $L_6$ & -- & 0.075 & 0.102 & 0.110 & 0.148 \\
GaAs & III--V & $m_{n,\parallel}$ & $L_6$ & -- & 1.900 & 1.660 & 1.660 & 1.188 \\
GaAs & III--V & $m_{p,\mathrm{hh}}$ & $\Gamma$ & [100] & 0.395 & 0.324 & 0.309 & 0.257 \\
GaAs & III--V & $m_{p,\mathrm{lh}}$ & $\Gamma$ & [100] & 0.090 & 0.034 & 0.077 & 0.060 \\
GaAs & III--V & $m_{p,\mathrm{so}}$ & $\Gamma$ & -- & 0.160 & 0.107 & 0.157 & 0.152 \\
GaN & III--V & $m_{n,\parallel}$ & $\Gamma$ & -- & 0.200 & 0.158 & 0.183 & 0.171 \\
GaN & III--V & $m_{n,\perp}$ & $\Gamma$ & -- & 0.200 & 0.175 & 0.202 & 0.186 \\
InP & III--V & $m_n$ & $\Gamma$ & -- & 0.079 & 0.054 & 0.086 & 0.075 \\
InP & III--V & $m_{p,\mathrm{hh}}$ & $\Gamma$ & [100] & 0.565 & 0.469 & 0.421 & 0.379 \\
InP & III--V & $m_{p,\mathrm{lh}}$ & $\Gamma$ & [100] & 0.120 & 0.073 & 0.114 & 0.098 \\
InP & III--V & $m_{p,\mathrm{so}}$ & $\Gamma$ & -- & 0.210 & 0.142 & 0.189 & 0.167 \\
CdS & Chalcogenides & $m_{n,\perp}$ & $\Gamma$ & A exciton & 0.192 & 0.133 & 0.174 & 0.162 \\
CdS & Chalcogenides & $m_{n,\parallel}$ & $\Gamma$ & A exciton & 0.168 & 0.123 & 0.159 & 0.152 \\
CdS & Chalcogenides & $m_{p,\perp}$ & $\Gamma$ & A exciton & 0.675 & 0.245 & 0.299 & 0.247 \\
CdS & Chalcogenides & $m_{p,\parallel}$ & $\Gamma$ & A exciton & 5.000 & 2.300 & 1.730 & 2.072 \\
CdTe & Chalcogenides & $m_{p,\mathrm{lh}}$ & $\Gamma$ & [100] & 0.130 & 0.058 & 0.104 & 0.073 \\
CdTe & Chalcogenides & $m_{p,\mathrm{hh}}$ & $\Gamma$ & [100] & 0.720 & 0.431 & 0.403 & 0.400 \\
MoS$_2$ & Chalcogenides & $m_p$ & $\bar{\Gamma}$ & $\bar{\Gamma}$--$\bar{K}$ & 0.670 & 0.775 & 0.649 & 0.709 \\
WS$_2$ & Chalcogenides & $m_p$ & $\bar{K}$ & $\bar{\Gamma}$--$\bar{K}$ & 0.350 & 0.354 & 0.300 & 0.322 \\
WS$_2$ & Chalcogenides & $m_{p,\mathrm{VB-1}}$ & $\bar{K}$ & $\bar{\Gamma}$--$\bar{K}$ & 0.430 & 0.495 & 0.431 & 0.447 \\
Si & Others & $m_{n,\perp}$ & CBM & -- & 0.191 & 0.193 & 0.186 & 0.243 \\
Si & Others & $m_{n,\parallel}$ & CBM & -- & 0.916 & 0.943 & 0.914 & 0.727 \\
Si & Others & $m_{p,\mathrm{hh}}$ & $\Gamma$ & [100] & 0.460 & 0.267 & 0.259 & 0.300 \\
Si & Others & $m_{p,\mathrm{lh}}$ & $\Gamma$ & [100] & 0.171 & 0.193 & 0.189 & 0.243 \\
Si & Others & $m_{p,\mathrm{so}}$ & $\Gamma$ & -- & 0.262 & 0.230 & 0.224 & 0.274 \\
SiC & Others & $m_{n,\parallel}$ & X & -- & 0.662 & 0.652 & 0.613 & 0.573 \\
SiC & Others & $m_{n,\perp}$ & X & -- & 0.244 & 0.225 & 0.218 & 0.247 \\
SiC & Others & $m_p$ & $\Gamma$ & [100] & 0.450 & 0.603 & 0.523 & 0.533 \\
BN & Others & $m_p$ & $\bar{K}$ & $\bar{\Gamma}$--$\bar{K}$ & 0.490 & 0.971 & 0.805 & 0.873 \\

\bottomrule
\end{longtable}
\normalsize

\clearpage
\section{\gzwz{} quasiparticle benchmark}
\label{sec:supp_gw}

The two-dimensional quasiparticle benchmark contains 286 nonmagnetic semiconductor records derived from C2DB, with eight valence states (VB8--VB1) and four conduction states (CB1--CB4) per material \citep{Haastrup2018C2DB,gjerding2021c2db,knosgaard2022gw_ml}. The target is the one-shot \gzwz{} quasiparticle spectrum obtained from a GPAW PBE starting point \citep{Hedin1965GW,golze2019gw_review,Mortensen2024GPAW}. Training and validation use the available high-symmetry vertices together with up to nine evenly spaced stored interior points per high-symmetry path segment. The \gzwz{} task is trained separately from the three-dimensional HSE06 model while retaining the same state-resolved correction framework.

\begin{figure}[!htbp]
\centering
\includegraphics[width=\textwidth]{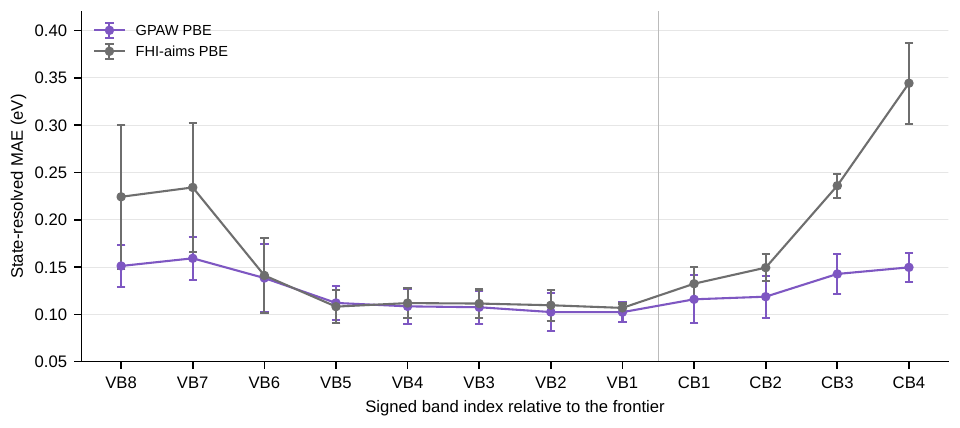}
\caption{
\textbf{State-resolved $G_0W_0$ quasiparticle prediction errors.}
Mean absolute error for individual valence (VB8--VB1) and conduction (CB1--CB4) states across the five material-held-out folds. The two variants differ only in the PBE eigenvalue baseline used to define the state-resolved correction target. The $G_0W_0$ labels, $\mathbf{k}$-point sets, and orbital inputs are otherwise identical.
}
\label{fig:s4}
\end{figure}

Five-fold material-level cross-validation is used with separate training, validation, and test folds. For each outer fold, three source folds are used for training, one fold is used for validation, and the remaining fold is evaluated as held-out test data. Performance is reported as the MAE over the twelve quasiparticle states and the corresponding band-gap MAE. The GPAW- and FHI-aims-baseline variants use identical \gzwz{} labels, $\mathbf{k}$-point sets, and real- and reciprocal-space orbital inputs \citep{Mortensen2024GPAW,Blum2009FHIaims}. Only the baseline PBE eigenvalues, and consequently the \gzwz{}--PBE correction targets, differ. The two variants are trained independently from scratch using otherwise matched model, sampling, and optimization settings. Supplementary Fig.~\ref{fig:s4} reports the state-resolved errors across the valence and conduction manifolds, and Supplementary Table~\ref{tab:s7} gives the fold-resolved results.

\begin{table}[!htbp]
\centering
\caption{
\textbf{Five-fold $G_0W_0$ quasiparticle benchmark.}
State and band-gap MAEs are reported in eV for independently trained models.
For each outer fold, three source folds train the model, one separate fold
selects the checkpoint by validation MAE, and the remaining fold is evaluated once as test data.
}
\label{tab:s7}
\small
\begin{tabular}{lrrrrrrr}
\toprule
Fold & Train & Validation & Test & GPAW state \citep{Mortensen2024GPAW} & GPAW gap & FHI-aims state \citep{Blum2009FHIaims} & FHI-aims gap \\
\midrule
0 & 171 & 57 & 58 & 0.1197 & 0.2132 & 0.1541 & 0.2374 \\
1 & 172 & 57 & 57 & 0.1380 & 0.2173 & 0.1811 & 0.2229 \\
2 & 172 & 57 & 57 & 0.1449 & 0.1995 & 0.1728 & 0.2018 \\
3 & 172 & 57 & 57 & 0.1237 & 0.1575 & 0.1645 & 0.2021 \\
4 & 171 & 58 & 57 & 0.1021 & 0.1396 & 0.1649 & 0.1654 \\
\midrule
Mean $\pm$ s.d.
& -- & -- & --
& $0.1257\pm0.0167$
& $0.1854\pm0.0349$
& $0.1675\pm0.0101$
& $0.2059\pm0.0272$ \\
\bottomrule
\end{tabular}
\end{table}

\clearpage
\section{Computational and reproducibility details}
\label{sec:supp_reproducibility}

Training of the three-dimensional models used dynamic batches of 1--12 band-structure entries, subject to a maximum of 700 atoms per batch and implementation-level padding limits for real-space and $\mathbf{k}$-resolved pairs. Mixed-precision computation used bfloat16 autocast. Training the final \ourmodel{} model requires approximately 6~min per epoch on a single NVIDIA H200 GPU, corresponding to approximately 10 GPU-hours over the 100-epoch training schedule. Reported three-dimensional evaluations use the raw \texttt{best\_avg\_win} checkpoint weights, selected by the lowest unweighted mean validation MAE over the 2, 5, 10, and 20~eV energy windows.

The seed-42 model contains 8,940,709 trainable parameters. The corresponding checkpoint, model configuration, parent-grouped split manifest, and seed-specific UID lists are retained with the reproducibility artifacts. Source code, model configurations, evaluation scripts, split manifests, and numerical figure-source data will be released with the public repository. The trained checkpoint will be distributed separately as part of the reproducibility package.

Available computational accounting for 6,576 source materials totals 88,051 CPU core-hours for PBE and 1,600,661 CPU core-hours for HSE06, corresponding to an aggregate HSE06-to-PBE computational-cost ratio of 18.18. These values represent cumulative CPU resource usage rather than wall-clock time and are not used to infer an end-to-end speedup.

\end{document}